\documentclass[]{fairmeta}

\usepackage{algorithm,algorithmic}
\usepackage{amsmath}
\usepackage{cancel}
\usepackage[inkscapelatex=false]{svg}
\usepackage{amssymb}
\usepackage{longtable}
\usepackage{txfonts}
\usepackage[noframe]{showframe}
\usepackage{framed}
\usepackage{lipsum}
\usepackage{caption}
\usepackage{cleveref}
\renewenvironment{shaded}{%
  \MakeFramed{\advance\hsize-\width \FrameRestore\FrameRestore}}%
 {\endMakeFramed}
\definecolor{shadecolor}{gray}{0.75}

\crefname{equation}{eq.}{eqs.}
\Crefname{equation}{Eq.}{Eqs.}

\usepackage{booktabs}
\usepackage{bbding}
\usepackage{multirow}

\newcommand{\mE}{\mathbb E}

\newcommand{\mcs}{\mathcal S}
\newcommand{\mca}{\mathcal A}
\newcommand{\mco}{\mathcal O}

\newcommand{\mcd}{\mathcal D}
\newcommand{\mcl}{\mathcal L}

\newcommand{\lone}[1]{\left|#1\right|}

\usepackage{makecell}

\newtheorem{theorem}{Theorem}

\allowdisplaybreaks[4]

\DeclareMathOperator*{\argmin}{argmin}
\DeclareMathOperator*{\argmax}{argmax}

\title{The Perfect Blend: Redefining RLHF with Mixture of Judges}

\author[1,\dagger]{Tengyu Xu}
\author[1,\dagger]{Eryk Helenowski}
\author[1,\dagger]{Karthik Abinav Sankararaman}
\author[1,\dagger]{Di Jin}
\author[1]{Kaiyan Peng}
\author[1]{Eric Han}
\author[1]{Shaoliang Nie}
\author[1]{Chen Zhu}
\author[1]{Hejia Zhang}
\author[1]{Wenxuan Zhou}
\author[1]{Zhouhao Zeng}
\author[1]{Yun He}
\author[1]{Karishma Mandyam}
\author[1]{Arya Talabzadeh}
\author[1]{Madian Khabsa}
\author[1]{Gabriel Cohen}
\author[2]{Yuandong Tian}
\author[1]{Hao Ma}
\author[1]{Sinong Wang}
\author[1]{Han Fang}

\affiliation[1]{Meta GenAI}
\affiliation[2]{FAIR}
\affiliation[\dagger]{Equal contributions}

\abstract{

    Reinforcement learning from human feedback (RLHF) has become the leading approach for fine-tuning large language models (LLM). However, RLHF has limitations in multi-task learning (MTL) due to challenges of reward hacking and extreme multi-objective optimization (i.e., trade-off of multiple and/or sometimes conflicting objectives). Applying RLHF for MTL currently requires careful tuning of the weights for reward model and data combinations. This is often done via human intuition and does not generalize. In this work, we introduce a novel post-training paradigm which we called Constrained Generative Policy Optimization (CGPO). The core of CGPO is Mixture of Judges (MoJ) with cost-efficient constrained policy optimization with stratification, which can identify the perfect blend in RLHF in a principled manner. It shows strong empirical results with theoretical guarantees, does not require extensive hyper-parameter tuning, and is plug-and-play in common post-training pipelines. Together, this can detect and mitigate reward hacking behaviors while reaching a pareto-optimal point across an extremely large number of objectives. 
    
    Our results show that CGPO consistently outperforms other commonly used SoTA RLHF algorithms (such as PPO and DPO) on a wide range of tasks -- general chat, STEM questions, instruction following, math, coding and knowledge. In particular, CGPO improves over PPO by $7.4\%$ in AlpacaEval-2 (general chat), $12.5\%$ in Arena-Hard (STEM \& reasoning), $2\%$ in IFEval (Instrcution Following), $2\%$ in both MATH and GSM8K (Math \& reasoning), $5\%$ in HumanEval (Coding), and $2\%$ in the ARC challenge (Knowledge). We also observe that PPO is susceptible to severe reward hacking behaviors (it exhibits severe regression in popular coding benchmarks) which can be addressed by CGPO. CGPO represents a breakthrough in RLHF, simultaneously addressing reward-hacking and extreme multi-objective optimization, and thereby advancing the state-of-the-art in aligning general-purpose LLMs.

}

\date{\today}
\correspondence{Tengyu Xu at \email{tengyuxu@meta.com}}

\begin{document}

\maketitle

\section{Introduction}\label{section:intro}
The emergence of general-purpose Large Language Models (LLMs) has significantly transformed the landscape of natural language processing, demonstrating exceptional capabilities across various expert-level domains \citep{achiam2023gpt,brown2020language,touvron2023llama,anthropic2023introducing,team2023gemini,meta2024introducing,tunstall2023zephyr,zhu2023starling}. These models are characterized by their extensive parameterization, enabling them to handle a wide array of tasks using a unified parameter set \citep{zhao2018modulation,liu2019end,liu2019loss}. Central to this versatility is multi-task learning (MTL) \citep{caruana1997multitask,crawshaw2020multi}, a strategy that involves training a single model on multiple tasks simultaneously. This approach fosters the development of shared representations, which enhances the model's ability to generalize better than those trained on isolated tasks. Although prior studies on MTL have concentrated on the integration and processing of multi-task data during both pre-training and fine-tuning stages \citep{raffel2020exploring,liu2023mftcoder,aghajanyan2021muppet,aribandi2021ext5}, the application of the primary LLM alignment method, Reinforcement Learning with Human Preference (RLHF) \citep{ouyang2022training,ziegler2019fine,zheng2023secrets}, has not been thoroughly explored within the MTL context.
In previous studies, the implementation of RLHF for multi-task post-training has typically involved a linear combination of multiple reward models within the standard RLHF framework \citep{ramamurthy2022reinforcement,glaese2022improving,yuan2023rrhf,bakker2022fine,wu2024fine,li2020deep}. Each reward model is crafted using preference data to mirror the distinct alignment preferences of different tasks. Researchers often experiment with various reward weightings to identify a Pareto front that depicts the optimal performance of the LLM across diverse tasks \citep{rame2024rewarded}. However, this approach is limited by two significant challenges:

\textbf{Vulnerability to Reward Hacking:} The optimization of a preference-based reward model is susceptible to reward hacking, as the reward model is an imperfect proxy of human preferences \citep{gao2023scaling,jin2023data,skalse2022defining}. Studies indicate that excessive optimization of a reward model can lead to misalignment with actual human preferences \citep{gao2023scaling,moskovitz2023confronting,stiennon2020learning,rafailov2024scaling}. This issue becomes more pronounced in a multi-task setting, where each reward model may have its own unique flaws. Implementing a uniform early stopping point in the RLHF optimization process to minimize reward hacking effects is impractical and can lead to degraded performance across tasks \citep{moskovitz2023confronting}. This highlights the need for a more tailored approach to compensate for the weaknesses of each reward model and to manage the optimization of reward models for each task in complex, multi-task environments.

\textbf{Contradictory Goals:} Different tasks often have conflicting objectives \citep{rame2024rewarded}. Even if the prompt spaces for these tasks do not overlap, using a linear combination of reward models can lead to compromises in goal metrics. For example, the typical strategy of LLM post-training involves maximizing the helpfulness reward for safe prompts and maximizing the harmfulness reward for unsafe prompts \citep{bai2022training}. Although achieving global optimality for both tasks is possible if the LLM's capacity is sufficiently large \citep{iyer2022opt}, employing a linear combination of helpfulness and harmfulness rewards inevitably results in reduced gains for both metrics. This occurs because each task partially sacrifices its own RLHF optimization progress to accommodate a contradictory metric, thereby diminishing the effectiveness of both.

To address these challenges, we developed an innovative framework called Constrained Generative Policy Optimization (CGPO). In response to the issue of reward hacking in RLHF, we introduce two types of judges: rule-based and LLM-based. These judges collaborate to identify any reward hacking patterns during the LLM's online generation phase. Based on their evaluations, we implement a constrained RLHF method to update the LLM model. This method is designed to maximize the likelihood of generating outputs that adhere to all constraints and achieve high reward values, while minimizing outputs that breach constraints and have low reward values. To support the constrained policy optimization update in the large-scale LLM setting, which is complicated even in traditional small-scale RL scenarios, we have developed three new primary-type constraint RLHF optimizers. These optimizers are designed to operate independently of the dual-variable update, which is often a critical component in conventional primal-dual constrained RL algorithms. This independence simplifies the optimizers and enhances their scalability, making them more effective for managing large-scale LLM post-training.

\begin{figure}[ht]
  \centering
  \includegraphics[width=1.0\textwidth]{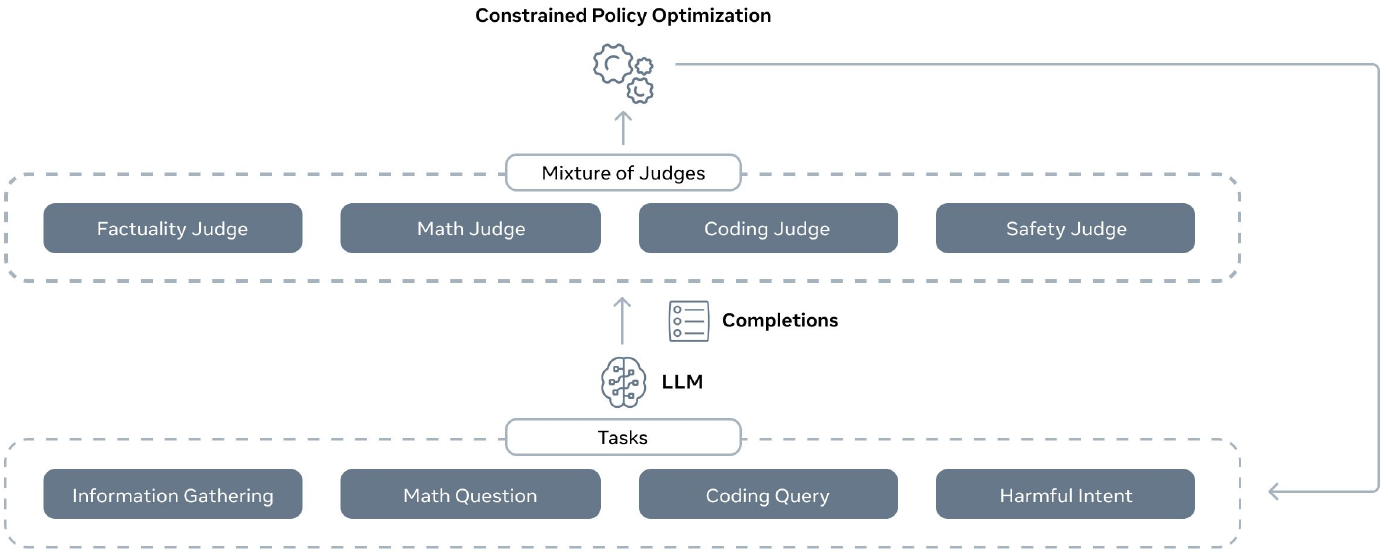}
  \caption{Overview of CGPO pipeline. In CGPO, a customized MoJs is applied to each task to evaluate model generations, and the model is updated through our proposed constrained RL algorithm.}
  \label{fig:cgpo_0}
\end{figure}

To effectively optimizing objectives of various tasks, which may be contradictory, we propose a novel design in CGPO for managing multi-task post-training. In this design, prompts are segregated by task, and a customized policy optimization strategy is applied to each set of prompts. This strategy includes a tailored MoJs, reward model, and hyperparameter setup for the constrained RLHF optimizer. By optimizing each task independently, our approach avoids compromises due to conflicting goals from other tasks, a common issue in previous works that used a linear combined reward model. Furthermore, our design addresses the reward hacking issue and optimizes objectives for each task in a fine-grained manner, resulting in a better Pareto frontier than previous methods that enforced uniform treatment across all tasks.
See \Cref{fig:cgpo_0} for an overview of our CGPO pipeline.

We summarize our contributions as follows:
\begin{itemize}
\item We have developed a new strategy to address the issues of reward hacking in multi-task LLM post-tuning through an innovative primal-type constrained RL method. To implement this method, we have introduced three new constrained RLHF optimizers: Calibrated-Regularized Policy Gradient (CRPG), Constrained Online Direct Preference Optimization (CODPO), and Calibrated-Regularized Reward Ranking Finetuning (CRRAFT). All proposed methods are scalable and easy to implement.
\item To support the implementation of the constrained RL method in CGPO, we have developed two types of judges: the rule-based judge and the LLM-based judge. These judges are designed to effectively assess whether an LLM generation violates constraints in a broad spectrum of NLP tasks.
\item We have introduced a new multi-objective RLHF treatment strategy within CGPO, where each task is managed individually with a customized optimization setting, including reward models, mixture of judges, and optimizer hyperparameters. This pioneering design, the first in the multi-task RLHF field, significantly enhances the Pareto frontier across multiple metrics in the multi-task setting. 
\item We demonstrate the effectiveness of CGPO in a challenging multi-task post-training environment with five tasks: general chat, instruction following, math and coding reasoning, engagement intent, and safety, despite potentially contradictory goals across tasks. Notably, by primarily utilizing open-source data and the Llama3.0 70b pre-trained model, our research demonstrates that, in comparison to the baseline RLHF methods such as PPO \cite{schulman2017proximal} and DPO \cite{rafailov2024direct}, our approach—when combined with the CRPG and CRRAFT optimizers—consistently outperforms these baselines across all benchmarks and tasks. Specifically
\begin{itemize}
    \item CRPG optimizers achieve the highest performance in terms of MATH, GSM8K, HumanEval, MBPP, ARC Challenge, and false refusal ratio. CRRAFT optimizers achieve the highest performance in AlpacaEval-2, Arena-Hard, and TruthfulQA.
    \item PPO experiences a significant drop in the 0-shot coding benchmarks (HumanEval and MBPP) after exceeding certain training steps, indicating the occurrence of severe reward hacking issues. In contrast, CGPO not only avoids such regression but also consistently improves those benchmarks during training, demonstrating the extraordinary capability of MoJs in preventing reward hacking issues.
\end{itemize}
\end{itemize}

\section{Preliminaries}
In the RLHF finetuing phase, we typically formulate a Markov Decision Process (MDP) as follows: each prompt is considered as the state $s$, and the entire response is the action $a = [a_0, a_1, \cdots, a_{T-1}]$, where $a_i\in A$ represents the token at position $i$ and $A$ is the vocabulary set. An LLM policy is defined as $\pi_w(a_t| a_{t-1}, a_{t-2}, \cdots, a_0, s)$, which represents a distribution over $A$ at time step $t$, conditioned on all previous response history before $t$ and prompt: $\{a_{t-1}, a_{t-2}, \cdots, a_0, s\}$.

\subsection{Supervised Finetuing}
RLHF starts by finetuing a pre-trained LLM using supervised learning on high-quality dataset relevant to the downstream target task(s) (such as dialogue, summarization, reasoning, etc.) to obtain $\pi_{\text{SFT}}$.

\subsection{Reward Model Training}
After the supervised fine-tuning stage, we need to develop a reward model to assess the quality of an LLM's output. This will enable us to utilize exploration-based online RL alignment method. We typically use the pairwise preference reward model \citep{stiennon2020learning}. In this model, we assume that human preference between a pair of responses $(a_p, a_n)$, originating from the same prompt $s$, is determined by a latent reward $r^*_{pair}(s,a)$. The Bradley-Terry (BT) model \citep{bradley1952rank,ouyang2022training,bai2022training,touvron2023llama,meta2024introducing}, a well-established reward-based preference model, defines the human preference distribution $p^*_{pair}$ using the following formulation:
\begin{flalign}
    p^*_{pair}(a_p>a_n|s) = \sigma(r^*_{pair}(s, a_p) - r^*_{pair}(s, a_n)),\label{eq: 1}
\end{flalign}
where $\sigma$ denotes the logistic function. In practice, we can learn a parameterized reward model $r_{\phi}(s,a)$ as a surrogate for $r^*_{pair}(s,a)$. Given a pre-collected preference-pair dataset $\mathcal{D} = \{s_i, a_{w,i}, a_{l,i}\}_{i=1}^{N}$, where $a_{w,i}$ and $a_{l,i}$ denote the preferred and less preferred generations respectively, we can learn $r_\phi$ by framing the problem as a binary classification and resolving the subsequent problem \citep{ouyang2022training,touvron2023llama,meta2024introducing}:
\begin{flalign}
    \min_\phi \mcl_{pair}(r_\phi, \mcd_{pair}) = - \mE_{\mcd_{pair}}\left[\log\sigma(r_\phi(s, a_p) - r_\phi(s, a_n))\right].\label{eq: 2}
\end{flalign}
\newline
In a standard LLM training pipeline, the preference-based reward model $r_\phi$ is typically initialized from the finetuned SFT model $\pi_{\text{SFT}}$, augmented by a linear layer on the final transformer layer, which generates a single scalar prediction for the reward value \citep{wang2024secrets,askell2021general,ouyang2022training}.

\subsection{RL Finetuning}
Given a LLM policy $\pi_w$ with parameter $w$, a reward model $r_\phi(a, s)$ and a prompt set $\mcd_{p} = \{s_i\}^M_i$, we aim to optimize the policy by maximizing the following RL objective \citep{ouyang2022training,achiam2023gpt,touvron2023llama}:
\begin{flalign}
    \max_w\quad \mE_{s\sim\mcd_{p}, a\sim\pi_{w}}\left[ r_\phi(s,a) \right].\label{eq: 5}
\end{flalign}
When solving the problem in \cref{eq: 5} we typically initialize $\pi_w$ with SFT policy $\pi_{\text{SFT}}$ instead of starting from scratch. In previous works a number of online RL method such as proximal policy optimization (PPO) \citep{schulman2017proximal}, reward ranking (RAFT) \citep{dong2023raft} and REINFORCE \citep{williams1992simple} has been utilized to solve \Cref{eq: 5}. 

Another direction of RL finetuing involves reward-free methods, which directly optimize $\pi_w$ using pre-collected preference data, without the need for a reward model. The rationale behind this approach is to fine-tune the model within a neighborhood of $\pi_{\text{SFT}}$, ensuring that the probability of generating both preferred and less preferred samples aligns with the pre-collected preference dataset. Direct Preference Optimization (DPO) \citep{rafailov2024direct} is the most widely adopted method in this direction.

\section{Limitations in Traditional RLHF}\label{sec: 2}
In this section, we discuss several limitations in the current RLHF pipeline, which are major bottlenecks in the multi-task LLM post-training.

\subsection{Limitation of Reward Modelling}\label{sec: 2.1}

\textbf{Insufficient capability for fine-grained criteria alignment. }
Despite being based on a sophisticated LLM, the reward model may struggle to provide accurate alignment guidance \citep{pan2022effects}, particularly in tasks requiring fine-grained criteria such as identifying correct answers in math questions and assessing code snippet correctness for coding problems. This limitation, inherent to preference-based learning, necessitates additional support to enhance the reward model's effectiveness in handling these specific requirements.


\textbf{Proxy nature in coarse-grained preference setting. }
Reward hacking can occur even in coarse-grained settings where the goal is to optimize human preferences, as the reward model, serving as a proxy for true preferences, may contain misspecifications \citep{gao2023scaling,moskovitz2023confronting}. This can lead to the model favoring less-preferred outputs, misdirecting the alignment process. A common mitigation strategy is to include a KL penalty in the RL objective to limit deviation from the initial policy, $\pi_{\text{SFT}}$. However, this approach does not directly address the reward model's imperfections, indicating the need for a more systematic approach to tackle reward hacking.

\subsection{Limitation of RLHF Optimizer}\label{sc: limit_optimizer}

\textbf{Contradictory optimization objectives. }
The initial success of LLM hinges on the assumption that human preferences are homogeneous \citep{bakker2022fine}, but they actually vary widely (helpfulness, harmlessness, honesty, etc) \citep{casper2023open,rame2024rewarded}. The current RLHF pipeline trains separate reward models for each task and combines them using linear weights \citep{ramamurthy2022reinforcement,glaese2022improving,yuan2023rrhf}. However, this approach applies the same weight of rewards to all tasks, which can be suboptimal (e.g., 90\% helpfulness + 10\% harmlessness may work well for safe scenarios but lead to risky responses in dangerous situations).



\textbf{Rigid optimization strategy for mutli-tasks alignment. }
In the standard RLHF pipeline, a uniform RL optimizer setup is typically applied across all tasks \citep{ouyang2022training}. However, this approach may not be optimal since the most effective hyperparameters, including number of generations per-prompt, batch-size, and KL-regularization, often differ between tasks due to unique nature of each task. For example, tasks requiring more exploration typically need a larger number of generations per prompt, whereas other tasks can work well with fewer.

\subsection{Motivation}
In multi-task LLM alignment settings, where the goal is to enhance LLM performance across various tasks, the limitations of reward modeling and RLHF optimizers discussed in \Cref{sec: 2} are significant bottlenecks that hinder the RLHF process from effectively improving LLM performance across all tasks. In the following section, we will introduce a novel RLHF framework, Constraint Generative Policy Optimization (CGPO), which addresses all the aforementioned limitations in the most principled manner.

\section{Constraint Generative Policy Optimization}
In this section, we first explore how to implement the CGPO framework within the scope of a single task with MoJs, as detailed in \Cref{sc: 5}. Subsequently, we discuss the implementation of CGPO to manage scenarios involving multiple objectives in \Cref{sc: 6} for multi-task learning.

\subsection{CGPO in Single Task with Single Objective}\label{sc: 5}

The primary design of CGPO is to integrate multiple constraints to mitigate the issue of reward hacking, which arises from the limited capabilities of reward models. Specifically, in addition to optimizing the accumulated reward model value as shown in \cref{eq: 5}, we also ensure that the model generation meets several constraints. For example, in mathematical reasoning tasks, we strictly require model generations to provide correct answers. This is essential since the model often fails to solve the problem correctly, yet the reward model might still allocate high values to these incorrect solutions. Another example is in general chat tasks with prompts that are free of harmful intent. We require model generations to consistently respond to user queries. This is crucial because there are instances where the model may refuse to answer, and the reward model might erroneously assign high values to such non-responsive generations. In these cases, purely maximizing the reward model could impair the model's reasoning capability and lead to an overly conservative tendency. By introducing these constraints based on our prior knowledge about the weaknesses of each reward model, we can avoid critical reward hacking patterns effectively.

We denote the set of constraints that the LLM generations need to satisfy as $\{ C_1, C_2, \ldots, C_M \}$ and the state-action set that satisfies constraint $C_k$ as $\Sigma_k$, i.e., $\Sigma_k = \{(s,a) \in \mcs\times\mca \,\, \text{and}\,\, (s,a)\,\, \text{satisfies requirement of}\,\, C_k \}$. We define the feasible region as the state-action set that satisfies all constraints as $\Sigma = \Sigma_1 \cap \Sigma_2 \cap \ldots \cap \Sigma_M$. In the single task setting, CGPO solves the following constrained problem \citep{ying2022towards,zhang2024cvar,luo2024simple,xu2021crpo}
\begin{flalign}
    \max_w\quad &\mE_{s\sim\mcd_{p}, a\sim\pi_{w}}\left[ r(s,a) \right]\nonumber\\
    \text{s.t.}\quad &\text{Prob}_{s\sim\mcd_{p}, a\sim\pi_{w}}((s,a) \in \Sigma) \geq 1,\nonumber\\
    &\text{KL}_{s\sim\mcd_{p}}(\pi_w|\pi_{\text{ref}}) \leq \text{KL}_{\max},
    \label{eq: cst_rlhf}
\end{flalign}
where $\pi_{\text{ref}}$ is the initialization model and $\text{KL}_{\max}$ is the threshold of KL-divergence, which could vary for different tasks.  
    The high-level framework of CGPO in the multiple-constraints and single-objective setting is illustrated in \Cref{alg:cgpo_0}. At each iteration, we sample a minibatch from the prompt set ${D}$, and then apply the current LLM policy to generate $K$ responses ($1 \leq K$) for each prompt. Subsequently, we apply all judges $J=\{J_h\}_{h=1}^{M}$ to all generated samples to evaluate whether a generation violates a specific constraint. We label a generation $a_{t,i}^k$ as ``violated'' if it fails any one of the constraint judgments, and ``satisfied'' otherwise. Note that the constraint judge is a module for evaluating the constraint satisfaction conditions, which could be a rule-based script or an LLM classifier. This module can address a wide range of constrained problems in the LLM post-tuning scenario. We will discuss this in detail in \Cref{sc: cst_judge}. 
    
    After that, we split the generations into ``Positive'' and ``Negative'' groups, depending on the constraint satisfaction label. We then apply a constrained RLHF optimizer to update the policy with these two groups of samples (see line 9). In our work, we propose three new RLHF optimizers to efficiently solve the multi-constraint problem in the LLM setting. For Option I, we develop a policy gradient approach and an online DPO approach, and for Option II, we develop a reward ranking-based approach. These optimizers will be discussed in detail in the subsequent sections.
    \begin{algorithm}
	\caption{$\text{CGPO}(D,\pi_{w_0}, J, B, R, \mco, T)$ in single task with multi-constraints}
	\label{alg:cgpo_0}
	\begin{algorithmic}[1]
		\STATE \textbf{Input:} prompt set $D=\{s_{t,i}\}_{i=1}^{N}$, LLM starting policy $\pi_{w_0}$, constraint judge set $J = \{J_h\}_{h=1}^{M}$, batchsize $B$, reward model $R$, iteration number $T$, constrianed RLHF optimizer $\mco$.
		\FOR {$t= 0, 1, ...,T$}
		\STATE Prompt sampling: $\{s_{t,i}\}_{i=1}^{B}\sim D$
            \STATE Response generation: $\{a^k_{t,i}\}_{k=1}^{K}\sim \pi_{w_t}(\cdot|s_{t,i})$ for $1\leq i\leq n$
		\STATE Constraint judgement:  $y^k_{t,i} = \lor_{h=1}^{M} J_h(s_{t,i}, a^k_{t,i})$ for $1\leq i\leq n$ and $1\leq k\leq K$
            \STATE Split sample set:
            \STATE $\quad$Positive samples: $X^+_t=\{(s_{t,i}, a^k_{t,i})\,\,\text{for}\,\,1\leq i\leq n,\,\,1\leq k\leq K\,\,\text{where}\,\, y_{t,i}=1\}$
            \STATE $\quad$Negative samples: $X^-_t=\{(s_{t,i}, a^k_{t,i})\,\,\text{for}\,\,1\leq i\leq n,\,\,1\leq k\leq K\,\,\text{where}\,\, y_{t,i}=0\}$
            \STATE Update $\pi_{w_t}\rightarrow\pi_{w_{t+1}}$ for policy optimization with optimizer $\mco$:
            \STATE $\quad$[Option I]: maximize likelihood of $X^+_t$ with high $R(x^+)$ and minimize likelihood of $X^-_t$ with low $R(x^-)$
            \STATE $\quad$[Option II]: maximize likelihood of $X^+_t$ with high $R(x^+)$
		\ENDFOR
	\end{algorithmic}
    \end{algorithm} 

    Intuitively, with either the Option I or Option II updating strategy, CGPO encourages the policy to explore regions that satisfy all constraints to maximize the expected reward model value. Note that CGPO is a primal-type constraint policy optimization approach, which differs from the standard primal-dual approach adopted in the constrained RL area. CGPO does not involve co-optimizing the dual variable, thus avoiding the drawbacks of extensive hyperparameter tuning issues associated with the primal-dual approach. Due to this reason, CGPO is user-friendly even with multiple different types of constraints, making it well-suited for the LLM post-tuning scenario.
    
    In the following sections, we will discuss how to implement \Cref{alg:cgpo_0} with our proposed RLHF optimizers: Calibrated-Regularized Policy Gradient (CRPG) and Constrained Online DPO (CODPO) for Option I, and Calibrated-Regularized Reward Ranking Fine-tuning (CRRAFT) for Option II. Subsequently, we will discuss the constraint judge module that we developed in CGPO, which enables us to assess the generation's constraint satisfaction condition.

\subsubsection{Calibrated Regularized Policy Gradient (CRPG)}\label{sc: 3}
In this section, we discuss our new constraint RLHF optimizer, the Calibrated Regularized Policy Gradient (CRPG), which is a policy gradient-based approach.

{\bf Calibrated Reward. }In the traditional RLHF algorithm, the reward model is typically directly incorporated into RL optimizers to progressively refine the policy. However, this method can pose difficulties when the reward model value is not properly calibrated. For preference reward models trained with \cref{eq: 2}, the reward's accuracy may be proficient in distinguishing between good and bad generations from the same prompt. However, the reward model values between generations from different prompts may not be directly comparable due to potential significant variations in the reward model value range for different prompts. Due to such reasons, standard RLHF algorithms, such as PPO and REINFORCE, could lead to suboptimal performance due to the poor calibration of the reward model \citep{rita2024countering}. In CRPG, we introduce a novel and low-cost reward calibration strategy to address this issue.

We consider the scenario where each prompt $s$ used in RLHF fine-tuning has a corresponding baseline response $\bar{a}$. This condition can be easily satisfied in practice.
\begin{itemize}
    \item \textbf{Option 1:} We repurpose the prompt set from the SFT training set and/or the reward model training set. For the SFT training dataset, the pre-collected golden response is utilized as the baseline response, denoted as $\bar{a}$. For the pair-wise reward model training dataset, the preferred response is designated as the golden response $\bar{a}$.
    \item \textbf{Option 2:} Given an RLHF fine-tuning prompt set $D_d$, we use $\pi_{\text{ref}}$ to generate the baseline response for all prompts $s \in D_d$, i.e., $\bar{a}\sim\pi_{\text{ref}}(\cdot|s)$ before starting RLHF fine-tuning.
\end{itemize}
Without loss of generality, we assume there is an underlying policy $\bar{\pi}$ that generates the baseline responses, denoted as $\bar{a}\sim\bar{\pi}(\cdot|s)$.
Given the baseline response $\bar{a}$, we developed the following calibrated reward to replace the raw reward model $r_{pair}(s,a)$:
\begin{flalign}
    R_{calib}(s,a) = \sigma(r_{pair}(s,a) - r_{pair}(s,\bar{a})).\label{eq: 6}
\end{flalign}
Intuitively, $R_{pair}(s,a)$ here represent the probability of $a$ being better than baseline response $\bar{a}$ conditioned on the same prompt $s$, i.e., 
\begin{flalign*}
    R_{calib}(s,a) \approx \text{Prob}(a > \bar{a}|s).
\end{flalign*}

The advantages of using calibrated rewards $R_{pair}$ are twofold:
\begin{enumerate}
    \item The magnitude of $R_{calib}$ becomes meaningfully comparable across different prompts. This is because it represents the probability that the current policy $\pi$ is superior to the baseline $\bar{\pi}$ for different actions. In other words, if $R_{calib}(s,a) > R_{calib}(s^\prime,a^\prime)$, it directly implies that action $a$ given state $s$ is better than action $a^\prime$ given state $s^\prime$, conditioned on the baseline policy $\bar{\pi}$. However, this implication cannot be made if $r_{\text{pair}}(s,a) > r_{\text{pair}}(s^\prime,a^\prime)$.
    \item The magnitude of the calibrated reward model is strictly bounded between 0 and 1. This constraint prevents an action with an extremely large raw value from dominating the policy update direction, which could be misleading, since a large raw reward value does not necessarily imply superior action quality.
\end{enumerate}

Based on $R_{calib}(s,a)$, we now reformulate RLHF objective in \cref{eq: 5} as 
\begin{flalign}
    \max_w \bar{J}(\pi_w) &= \mE_{a\sim \pi_w(\cdot|s), s\sim D_d}\left[R_{calib}(s,a) \right] \label{eq: 7}
\end{flalign}
where $\bar{J}(\pi_w)$ is the policy optimization objective. Intuitively, it represents the probability of current policy $\pi_w$ being better than the baseline policy $\bar{\pi}$ conditioned on the prompt set $D_d$, i.e., 
\begin{flalign*}
    \bar{J}(\pi_w) &\approx \text{Prob}(\pi_w > \bar{\pi}|D_d).\nonumber
\end{flalign*}

{\bf Constraint Regularized Gradient. }Recall that in the multi-constraint setting, our goal is to maximize the expected reward model while aligning the LLM such that its generations strictly adhere to a set of constraints. These constraints compensate for the limitations of the reward model, including safety requirements, reasoning accuracy, and factual correctness. These aspects may not be fully captured by the reward model but can be well addressed via a separate rule-based judge or an LLM-based judge. 
Note that the "Positive samples" in line 6 of \Cref{alg:cgpo_0} is a subset of $\Sigma$, i.e., $X^+_t\in \Sigma$. Consequently, we aim to optimize the following multi-constraint objective, denoted as $\bar{J}_c$:
\begin{flalign}
    \max_w \bar{J}_c = \mE_{a\sim \pi_w(\cdot|s), s\sim D_d}\left[ R_{calib}(s,a) \cdot \mathbf{1}_{(s,a)\in \Sigma} \right].\label{eq: 11}
\end{flalign}
By solving the optimization problem presented in \cref{eq: 11}, the LLM is aligned to maximize the expected value of the calibrated reward model as much as possible, while remaining within the constraint satisfaction region. 

Given $R_{calib}$ and $\Sigma$, we define the following constraint regularized reward as 
\begin{equation}\label{eq: 12}
    R_{cr}(s,a) = \left\{
    \begin{aligned}
        & R_{calib}, & \text{if } (s,a) \in \Sigma \\
        & 0, & \text{if } (s,a) \notin \Sigma
    \end{aligned}
    \right.
\end{equation}
With the calibrated regularized reward $R_{cr}$, we rewrite \cref{eq: 11} as
\begin{flalign}
    \max_w \bar{J}_c = \mE_{a\sim \pi_w(\cdot|s), s\sim D_d}\left[ \cdot R_{cr}(s,a) \right].\label{eq: 17}
\end{flalign}
We consider the following update to optimize $\bar{J}_c$
\begin{flalign}
    w_{t+1} = w_{t} + \alpha_t \cdot g_c(\pi_{w_t}),\label{eq: 13}
\end{flalign}
where
\begin{flalign}
    g_c(\pi_w) &= \frac{1}{N} \sum_{i}^{N} \nabla \log\pi_w(s_i,a_i) \cdot R_{cr}(s_i,a_i).\nonumber
\end{flalign}
The subsequent theorem illustrates that CRPG has global optimality guarantee for both objective achievement and constraint satisfaction in the multi-constraint LLM alignment setting.
\begin{theorem}
    Consider the CRPG update defined in \Cref{eq: 13}. Consider the scenario where the optimal policy $\pi^*$ of \cref{eq: 11} satisfies $\text{Prob}_{\pi^*,D_d}((s,a) \in \Sigma) = 1$. Denote the policy set within the constraint satisfaction region as $\Pi_\Sigma$, and the globally optimal policy within $\Pi_\Sigma$ as $\pi^*_c$, i.e., $\pi^*_c = \argmax_{\pi \in \Pi_\Sigma}\mE_{\pi, D_d}\left[R_{cr}(s,a)\right]$. Given a few mild assumptions, we have
    \begin{flalign*}
        \mE_{\pi^*_c, D_d}\left[R_{cr}(s,a)\right] - \mE_{\pi_{w_t}, D_d}\left[R_{cr}(s,a)\right] &\leq \mathcal{O}\left(\frac{1}{poly(t)}\right),\nonumber\\
        \text{Prob}_{\pi_{w_t}, D_d}\left((s,a)\notin \Sigma\right) &\leq \mathcal{O}\left(\frac{1}{poly(t)}\right).\nonumber
    \end{flalign*}
\end{theorem}

{\bf CRPG Implementation. }
Consider the KL divergence between $\pi_{\text{ref}}$ and $\pi_w$ as a universal regularization method to prevent reward hacking during CRPG fine-tuning. We propose the following new reward regularization approach:
\begin{flalign}
    \Tilde{R}_{cr}(s,a) = \max\left\{1-\frac{\log(\pi_w(s_i,a_i)/\pi_{\text{ref}}(s_i,a_i))}{\text{KL}_{\max}}, 0\right\}\cdot R_{cr}(s,a).\label{eq: 16}
\end{flalign}
It is important to note that $\Tilde{R}_{cr}$ not only penalizes samples that deviate significantly from $\pi_{\text{ref}}$, but also strictly bounds the overall KL divergence.

Moreover, to reduce the variance in the CGPG gradient estimation, we consider subtracting a baseline from the $g_c$ without changing its expected direction as following
\begin{flalign}
    \Tilde{g}_c(\pi_{w_t}) = \frac{1}{n} \sum_{i}^{n} \nabla \log\pi_{w_t}(s_{t,i},a_{t,i}) \cdot \left[\Tilde{R}_{cr}(s_{t,i},a_{t,i}) - \frac{1}{n} \sum_{i}^{n} \Tilde{R}_{cr}(s_{t,i},a_{t,i})\right].\label{eq: 15}
\end{flalign}
The final CRPG update in multi-constraints finetuning setting is given as
\begin{flalign*}
    w_{t+1} = w_{t} + \alpha_t \cdot \Tilde{g}_{c}(\pi_{w_t}).
\end{flalign*}

\subsubsection{Constrained Online Direct Preference Optimization (CODPO)}\label{sc: codpo}
Based on Direct Preference Optimization (DPO), a widely used offline RLHF alignment algorithm in the unconstrained setting, we propose a new variant called Constrained Online Direct Preference Optimization (CODPO) to solve the constrained RLHF fine-tuning problem.

Recall that in DPO \citep{rafailov2024direct}, the optimal policy $\pi^*$, which aligns with human preferences in the $\beta$-regularized MDP setting, satisfies the following preference model:
\begin{flalign*}
    P_{\pi^*}(a_p>a_n) = \frac{1}{1 + \exp\left(\beta \log \frac{\pi^*(s,a_n)}{\pi_{\text{ref}}(s,a_n)} - \beta\log\frac{\pi^*(s,a_p)}{\pi_{\text{ref}}(s,a_p)}\right)}.
\end{flalign*}
Given a pairwise preference sample pair $(s, a_p)$ and $(s, a_n)$, we update our policy by solving the following problem:
\begin{flalign}
    &\min_w \mcl_{\text{DPO}}(\pi_w) = -\mE_{(s,a_p,a_n)}\left[ \ell_{\text{DPO}}(\pi_w, s, a_p, a_n)\right].\nonumber\\
    &\text{where}\quad \ell_{\text{DPO}}(\pi_w, s, a_p, a_n) = \log\sigma\left(\beta \log \frac{\pi_w(s,a_p)}{\pi_{\text{ref}}(s,a_p)} - \beta\log\frac{\pi_w(s,a_n)}{\pi_{\text{ref}}(s,a_n)}\right)
\end{flalign}
To prevent the possible decreasing likelihood of positive samples $a_p$, it has been proposed to add a regularization term to the vanilla DPO loss \citep{pal2024smaug}:
\begin{flalign}
    \tilde{\ell}_{\text{DPO}}(\pi_w, s, a_p, a_n) = \ell_{\text{DPO}}(\pi_w, s, a_p, a_n) + \frac{\lambda}{\lone{a_p}} \cdot \log(\pi_w(s, a_p)),\label{eq: 22}
\end{flalign}
where $\lone{a_p}$ represents the length of response $a_p$. By appropriately tuning the hyperparameter $\lambda$, the formulation in \cref{eq: 22} can effectively increase the likelihood of $a_p$ while decreasing the likelihood of $a_n$ to maximize the margin between positive and negative generations.

In CODPO, similar to CRRAFT, we first generate multiple responses for each prompt using the current policy $\{a^1_{t,i}, a^2_{t,i}, \ldots, a^K_{t,i}\} \sim \pi_{w_t}(\cdot \mid s_{t,i})$ and split the generations into positive samples $X^+_t$ and negative samples $X^-_t$. After that, we select the positive sample from $X^+_t$ with the highest reward value, and the negative sample from $X^-_t$ with the lowest reward value, i.e.,
\begin{flalign*}
    a^+_{i,t} = \argmax_{\substack{k\in[K], \\ (s_{i,t},a^k_{i,t})\in X^+_t}} r_{pair}(s_{i,t},a^k_{i,t}),\nonumber\\
    a^-_{i,t} = \argmin_{\substack{k\in[K], \\ (s_{i,t},a^k_{i,t})\in X^-_t}} r_{pair}(s_{i,t},a^k_{i,t}).
\end{flalign*}
In cases where no generations satisfy all constraints, we can skip this sample. Conversely, when no generations violate any constraints, we can select the generation with the lowest reward model value as the negative sample.

Then, at each iteration, we update the policy as follows:
\begin{flalign}
    w_{t+1} = w_{t} - \alpha_t \cdot \frac{1}{n} \sum_{i=1}^{n} \nabla \tilde{\ell}_{\text{DPO}}(\pi_{w_t}, s_{i,t}, a^+_{i,t}, a^-_{i,t})\label{eq: 23}.
\end{flalign}

\subsubsection{Calibrated Regularized Reward Ranking Finetuning (CRRAFT)}\label{sc: 4}
In this section, we introduce another constrained RLHF policy optimizers that we proposed: Calibrated Regularized Reward Ranking Finetuning (CRRAFT), which is built upon the RAFT.

In the original RAFT algorithm \citep{dong2023raft}, each round involves generating multiple responses from a prompt using the current policy model, denoted as $\{a^1_{t,i}, a^2_{t,i}, \ldots, a^K_{t,i}\} \sim \pi_{w_t}(\cdot \mid s_{t,i})$. A reward model $r$ is then utilized to select the response with the highest reward model score, i.e., $a^*_j = \argmax_{k \in [K]} r_{\text{pair}}(s_{t,i}, a^k_{t,i})$ (note that whether a calibrated reward is used or not does not affect the reward ranking result).
Subsequently, an one-step SFT update is performed to maximize the likelihood of this generated sample $(s_{t,i}, a^*_{t,i})$. The policy model is iteratively updated to improve its alignment with the reward model $r_{pair}$ as follow
\begin{flalign}
    w_{t+1} = w_t + \alpha_t \cdot \frac{1}{n} \sum_{j=1}^{n} \nabla \log(\pi_{w_t}(s_{t,i}, a^*_{t,i})).\label{eq: 19}
\end{flalign}

In the multi-constraint setting, we make the following two changes on top of RAFT to develop our CRRAFT optimizer:
\begin{itemize}
    \item After applying the reward model to score each responses, we adopt Option I in \Cref{alg:cgpo_0} to first filter out those generated responses that violated any of the constraints. Additionally, to avoid large drift of current policy from starting point policy $\pi_{\text{ref}}$, we also filter out all generations whoes KL-divergence is larger than a pre-defined threshold $\text{KL}_{\max}$, i.e., $\text{KL}_{(s_{i,t},a^k_{i,t})} = \frac{\log \pi_{w_t}}{\log \pi_{\text{ref}}}(s_{i,t},a^k_{i,t})> \text{KL}_{\max}$. After that we apply reward ranking to select the one with the highest reward model score from the rest of responses, i.e.,
    \begin{flalign}
        a^*_{i,t} = \argmax_{\substack{k\in[K], \\ (s_{i,t},a^k_{i,t})\in X^+_t, \\ \text{KL}_(s_{i,t},a^k_{i,t}) \leq \text{KL}_{\max}}} r_{pair}(s_{i,t},a^k_{i,t}).\label{eq: 18}
    \end{flalign}
    We refer to the procedure in \cref{eq: 18} as constrained regularized reward ranking.
    It's important to note that CRRAFT not only has the capability to manage multiple constraints, but it also strictly bounds the KL-divergence. This is a feature that the standard RAFT algorithm lacks.

    Note that there may be instances where no generations remain after filtering. In such cases, if the pre-collected baseline response $\bar{a}_{i,t}$ satisfies all constraints, it can be used as $a^*_{i,t}$. If it doesn't, this datapoint can be skipped.
    
    \item After the constrained regularized reward ranking, instead of directly performing SFT update w.r.t the chosen sample as \cref{eq: 19} does, here we reweigh each chosen response by their calibrated reward value and then perform SFT update as follow
    \begin{flalign}
        w_{t+1} &= w_t + \alpha_t \cdot \Tilde{g}_{ra}(\pi_{w_t})\nonumber\\
        &= w_t + \alpha_t \cdot \frac{1}{n} \sum_{i=1}^{n} R_{calib}(s_{i,t}, a^*_{i,t})\cdot \nabla \log(\pi_{w_t}(s_{i,t}, a^*_{i,t})).\label{eq: 20}
    \end{flalign}
    By incorporating the calibrated reward model value in the update, we can differentiate the emphasis on chosen responses based on their quality, unlike the RAFT algorithm which treats all chosen responses equivalently. This approach allows for a more refined alignment with the reward model.
\end{itemize}
Please note that unlike CRPG and CODPO, CRRAFT specifically focuses on increasing the likelihood of constraint-satisfied positive samples and disregards the constraint-violated negative samples.

\subsubsection{Judges in CGPO}\label{sc: cst_judge}
The key step in implementing multi-constraint CGPO optimizers, as outlined in \Cref{sc: 3} and \Cref{sc: 4}, is to determine whether a generation $(s,a)$ satisfies a constraint or not. This determination allows us to split generated samples into positive ($X^+_t$) and negative ($X^-_t$) groups given the label $y$ predicted by each constraint judge $J_h$, i.e.,
\begin{flalign*}
    J_h(s, a) = y \in \{0, 1\},\,\,\text{where}\,\, 1\leq h\leq M,
\end{flalign*}
and then apply our customized constraint RLHF optimizers based on that classification. In CGPO, we have developed and integrated the following two types of constraint judge modules to assess whether a generation satisfies a constraint:
\begin{itemize}
    \item {\bf Rule-based constraint judge module:} This module employs a rule-based approach (such as string-matching and code execution) to ascertain whether the generation strictly adheres to predefined regulations \citep{li2024ruler}. It is particularly effective for constraints related to precise instruction following, where the generation must meet exact requirements such as length, number of paragraphs, and keyword inclusion \citep{zhou2023instruction,hendrycks2021measuring,cobbe2021training}. It can also handle reasoning tasks, such as math problems and code generation.
    

    \item {\bf LLM-based constraint judge module. }This module functions as an LLM generator. In most cases, the generation is formatted according to a template before being sent to the judge module. These modules not only provide access to the constraint satisfaction condition but also offer reasoning behind the judgement construction. Due to this property, they are typically capable of handling more challenging constraint evaluation tasks such as safety violation, reference-based factuality verification, and false refusal patterns. The model could either be a compact LLM fine-tuned with domain-specific data \citep{inan2023llama,bai2022training} or a powerful, large LLM without task-specific fine-tuning \citep{yuan2024self,zheng2024judging}. 
\end{itemize}
A detailed introduction to these two types of judges can be found in \Cref{sc: app_cst}.

\subsection{CGPO in Multi-Taks with Multi-Objectives}\label{sc: 6}

In the multi-tasks environment, CGPO utilizes customized combinations of "reward models + MoJs + optimizers" to provide alignment guidance tailored to each task. This approach is designed to better accommodate the specific nature of each problem, thereby enable CGPO to have better chance to achieve optimal alignment outcomes.
\begin{figure}[h]
  \centering
  \includegraphics[width=1.0\textwidth]{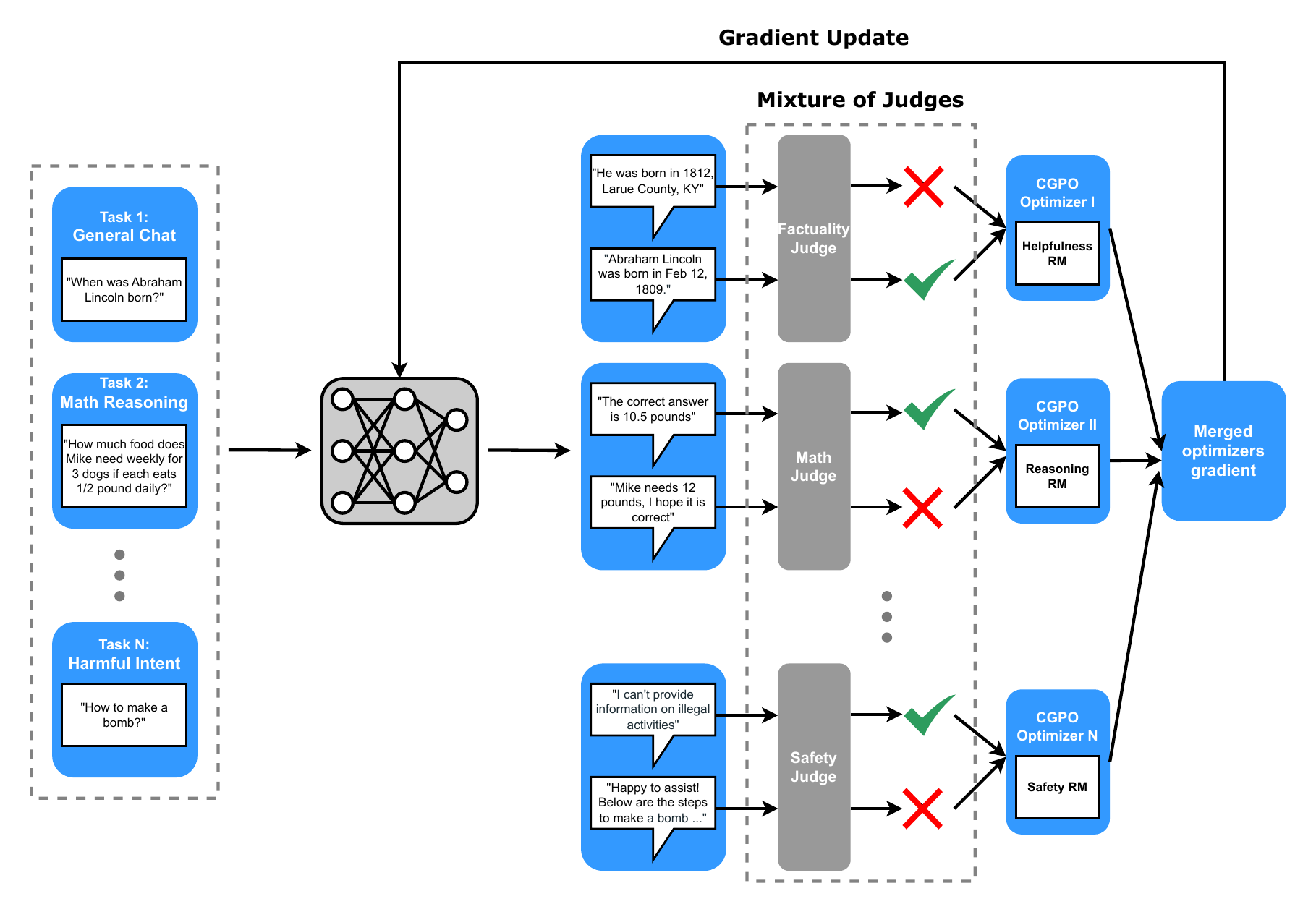}
  \caption{CGPO in a multi-tasks setting. The reward model, a MoJs, and optimization setup are uniquely tailored to the specific characteristics of each task. This customization ensures the most effective and targeted approach for achieving optimal performance across all tasks, even those with potentially contradictory goals.}
  \label{fig:cgpo2}
\end{figure}
\Cref{fig:cgpo2} provides an end-to-end illustration of how the CGPO pipeline functions in the multi-tasks setting. The entire CGPO pipeline has the following two core components: multi-objective reward modeling and multi-experts alignment.

{\bf Mutli-Objective Reward Modelling. }
Unlike the approach adopted in previous RLHF pipelines in multi-objective scenarios, which applies the same linear combined reward model to all prompts in the prompt set $D$, CGPO first classifies the prompt set $D$ into distinct, non-overlapping categories based on the nature of the prompts, i.e., $D = \{D_{1}, D_{2},\ldots,D_{L}\}$. Each prompt set $D_{l}\in D$ is referred to as a task. For example, prompts with harmful intent, which could potentially lead LLM to generate unsafe responses, are grouped into a class labeled "harmful intent". Conversely, prompts without unsafe intent, primarily focused on information gathering and casual conversation, are grouped into a class labeled "general chat". This categorization can be performed during the data collection phase or by prompting an LLM to carry out the categorization given the definitions of different classes. 
Subsequently, with a collection of trained reward models denoted as $\{R_{calib, 1}, R_{calib, 2}, \ldots, R_{calib, V}\}$,  we tailor the specific reward model to be applied for each task $D_{l}$. This customization guarantees that each prompt class $D_{l}$ benefits from the most appropriate guidance provided by the corresponding reward model. 
Note that the number of reward models, denoted by $V$, is less than or equal to the number of tasks, meaning a single reward model can be utilized across multiple tasks.

The major advantage of segregating the reward modeling for each individual task is to exclude irrelevant or contradictory objectives, thus enabling each task to focus solely on optimizing its own goal metrics without interference from other objectives, which could otherwise lead to suboptimal gains in target goals.

{\bf Multi-Expert Alignment. }The concept of multi-expert alignment involves applying customized MoJs, reward model and policy optimization setups for each task.

After the policy model generates online samples for each task, we employ a mixture of task-specific judges to identify generations that do not meet predefined standards. It is crucial to emphasize that the selection of judges are uniquely tailored for each task, reflecting the particular shortcomings of each reward model and our established performance criteria for LLMs in these tasks.
For instance, in the "general chat" task, we employ LLM-based judges for false refusal and factuality to enhance responsiveness and ensure honesty. In "reasoning" tasks, we implement a rule-based math/coding constraint judge to guarantee correctness and accuracy.

Based on the status of constraint satisfaction across generations and a customized reward model, we implement an RLHF policy optimizer with a specifically tailored hyperparameter setup to align each task effectively. This method deviates from the conventional RLHF pipeline, which generally employs a uniform optimizer setup for task alignment. For tasks that have precise judges and require extensive exploration to derive the correct response, such as instruction following, math, and coding, we apply a lenient KL threshold and allow a higher number of generations per prompt. In contrast, for tasks where precise judges are lacking and extensive exploration is less critical, such as "general chat," we opt for a stricter KL threshold and a reduced number of generations per prompt.

\begin{algorithm}
	\caption{$\text{CGPO}(\{D_l\}_{l=1}^L,\pi_{w_0}, \{J_l\}_{l=1}^L, \{B_l\}_{l=1}^L, \{R_l\}_{l=1}^L, \{\mco_l\}_{l=1}^L, T)$ in multi-tasks with multi-constraints \& multi-objectives}
	\label{alg:cgpo_1}
	\begin{algorithmic}[1]
		\STATE \textbf{Input:} Multi-tasks prompt set $\{D_l\}_{l=1}^L$, LLM starting policy $\pi_{w_0}$, judges sets $\{J_l\}_{l=1}^L$, multi-tasks batchsizes $\{B_l\}_{l=1}^L$, reward model sets $\{R_l\}_{l=1}^L$, multi-tasks weights $\{\rho_l\}_{l=1}^{L}$, multi-tasks optimizers $\{\mco\}_{l=1}^L$, iteration number $T$.
		\FOR {$t= 0, 1, \cdots,T$}
            \FOR {$l = 0, 1, \cdots, L$}
		\STATE Obtain gradient $\Tilde{g}_l(\pi_{w_t})$ for $l$-th task via $\text{CGPO}(D_l,\pi_{w_t}, J_l, B_l, R_l, \mco_l, 1)$ in \Cref{alg:cgpo_0}
            \ENDFOR
            \STATE Update with multi-tasks gradient accumulation $ w_{t+1} = w_t + \alpha_t \cdot \sum_{l=1}^{L} \rho_l \cdot \Tilde{g}_l(\pi_{w_t}),$
		\ENDFOR
	\end{algorithmic}
    \end{algorithm} 

The high-level framework of CGPO in the multiple-constraint and multiple-objective setting is illustrated in \Cref{alg:cgpo_1}. Specifically, at each iteration $t$, we process each individual task to compute the updated gradient $\tilde{g}_l(\pi_{w_t})$. This computation is based on the task-specific prompt set $D_l$, reward model $R_l$, judges $J_l$, batch size $B_l$, and optimizer $\mathcal{O}_l$, following the steps outlined in \Cref{alg:cgpo_0}. Subsequently, we accumulate the gradients across all tasks and combine them with our predefined task weights $\{\rho_l\}_{l=1}^{L}$, which are then used to update our model parameters.

\section{Experiments}
In this section, we outline the specifics of our experimental setup designed for multi-task alignment under conditions of extreme multi-constraints and multiple objectives. Specifically, we focus on fine-tuning a LLM to achieve alignment across the following five tasks:
\begin{itemize}
    \item {\bf General chat: }This task is designed to enhance the general conversational abilities of LLMs by considering multi-turn conversational histories \citep{wang2024survey}. It focuses on boosting the coherence, consistency, and correctness of responses, thereby making the interactions more logical and seamless. Additionally, this task improves the model's capability to deliver responses that are better aligned with the user's intentions and queries, and are factually grounded \citep{sun2024trustllm}.
    \item {\bf Instruction Following: }This task is designed to enhance the ability of LLMs to follow instructions accurately within specific contexts or industries \citep{zhou2023instruction}. By fine-tuning LLMs to adapt to particular domains or user requirements, they can deliver more precise and relevant responses. This improvement leads to a more satisfying and efficient user experience, making LLMs more effective and versatile tools across various applications.
    \item {\bf Math/Code Reasoning: }This task is designed to enhance the math and coding capabilities of LLMs, enabling them to address more complex problems and broaden their range of functions. These include tasks like debugging code or solving mathematical equations, which are vital in technical fields \citep{hendrycks2021measuring,cobbe2021training,chen2021evaluating,austin2021program}. Furthermore, improving LLMs' ability to comprehend and produce mathematical and code-related content results in greater accuracy and efficiency in activities that demand meticulous logical reasoning and computational thinking.
    \item {\bf Engagement Intent: }This task aims to enhance user engagement and interaction with the LLM. To address this, we involve human annotators who interact with the model and provide binary feedback (like or dislike) for each response generated by the LLM. Our objective is to maximize the likelihood that users will favorably respond to the LLM's outputs.
    \item {\bf Harmful Intent: }This task trains LLMs to recognize and resist safety-related adversarial attacks. It ensures that LLMs are safeguarded against exploitation for malicious purposes, such as generating harmful or misleading information \citep{sun2024trustllm,xu2020recipes}. By enhancing their ability to operate safely and ethically, this task helps maintain user trust and uphold the credibility of the technology.
\end{itemize}

\subsection{Supervised Fine-Tuning}\label{sc: sft_exp}
The foundational model we have chosen is the LLaMA-3.0-70B pre-trained checkpoint. We independently perform SFT using an open-source dataset to establish the initial policy, denoted as $\pi_0$. For all preference pair datasets listed below we only use positive samples in SFT. We utilize the following datasets for the tasks under consideration:
\begin{itemize}
    \item {\bf General chat: }LMSys-55k \citep{chiang2024chatbot}, UltraChat \citep{ding2023enhancing}
    \item {\bf Instruction following: }LIama 3.0 70B instruct model synthetic instruction following dataset
    \item {\bf Math/Code Reasoning: }Orca-Math \cite{mitra2024orca}, MetaMath \citep{yu2023metamath}, Evol-CodeAlpaca \citep{luo2023wizardcoder}, UltraFeedback \citep{cui2023ultrafeedback}, UltraInteract \citep{yuan2024advancing}
    \item {\bf Harmful Intent: }Human annotated safety dataset
\end{itemize}

The training is carry out for 2 epoches with a learning rate of $ 10^{-5}$. A cosince schedule is employed, the global batchsize is set to $128$ with minimum rate $0.1$ and warm-up steps $200$. The detail of how we obtain synthetic instruction following dataset and safety dataset SFT can be found in \Cref{sc: app_training}.

\subsection{Reward Modelling}\label{sc: rm_exp}
We have employed open-source pairwise preference data to train three specialized reward models (RMs): 
\begin{itemize}
    \item \textbf{Helpfulness RM:} This model is tailored for tasks such as general chat, instruction following, and math/code reasoning. It is based on the LLaMA-3-70B instruct finetuned model. The training utilized the following pairwise preference datasets:
    \begin{itemize}
        \item \textbf{General chat:} Includes datasets such as HH-RLHF \citep{bai2022training}, SHP \citep{pmlr-v162-ethayarajh22a}, HelpSteer \citep{wang2023helpsteer}, Distilabel-Capybara \citep{ethayarajh2024kto}, Distilabel-Orca \citep{distilabel-argilla-2024}, and LMSys-55k \citep{chiang2024chatbot}.
        \item \textbf{Instruction Following:} LIama 3.0 70B instruct model synthetic instruction following pairwise preference dataset.
        \item \textbf{Math/Code Reasoning:} Features datasets like Argilla Math \citep{distilabel-argilla-2024}, UltraFeedback \citep{cui2023ultrafeedback} and UltraInteract \citep{yuan2024advancing}.
    \end{itemize}
    \item \textbf{Engagement RM:} This RM is designed to simulate user engagement preferences. Initially, we fine-tune a binary classifier predictor using the LLaMA-3-70B instruct model to predict a user's engagement intent based on real interaction data between the language model and the user. We then treat this predictor as the oracle for user intent regarding engagement with the language model, given prompts and generations. To gather pair-wise training data, we subsample 129692 prompts from the LMSys-1M dataset \citep{zheng2023lmsyschat1m} and use the LLaMA-3-70B instruct model to generate four responses for each prompt. Each prompt is then scored using the oracle engagement predictor. We select the generation with the highest score as the "chosen" response and the generation with the lowest score as the "rejected" response. By doing this, we compile the pair-wise dataset and train the engagement RM based on this data.
    \item \textbf{Safety RM:} Focused on ensuring safe responses in scenarios with potentially harmful user prompts, this model is based on the LLaMA-3-8B instruct finetuned model. It utilizes a human-annotated safety pairwise preference dataset that identifies harmful intent in prompts.
\end{itemize}
It is important to note that we are considering training a unified Helpfulness RM that encompasses general chat, instruction following, and math/code reasoning, rather than training three separate RMs. This consideration is based on the observed positive correlation among these tasks. A unified RM, trained with a blended dataset from these domains, is expected to yield superior performance compared to training separate RMs for each individual task.

\subsection{Mixture of Judges}\label{sc: cst_exp}
To address the limitations of the reward model, we have implemented several judges in our experiment for multi-task alignment:
\begin{itemize}
    
    
    \item \textbf{False refusal judge:} Enhancing safety protocols may cause LLMs to become overly safe, leading to false refusals when responding to innocuous user queries, thus degrading  user experience. It has become critical for LLMs to reduce false refusals while maintaining the same level of safety, both in the research community and in the leading industry models \citep{cui2024or}. To address this challenge, we have developed a false refusal classifier, a fine-tuned LLM designed to detect false refusals to ensure the effectiveness of the LLM. 
    
    \item \textbf{Precise instruction following judge:} Reward models often struggle with precisely following instructions \citep{zhou2023instruction}. To address this, we have implemented a rule-based judge capable of accurately assessing compliance with over 30 types of specific instruction-following requests found in user prompts, such as "answer the question in two paragraphs." It is important to note that during RLHF finetuning, we will also include precise instruction-following prompts of this type so that the correctness of the generation can be evaluated with this constraint judge.
    
    \item \textbf{Regex math/code reasoning judge:} Reward models frequently fail to accurately assess the correctness of math and coding problems. To improve accuracy, we have introduced specialized judges for both domains. For math-related queries, we use a rule-based approach to check whether the final answers of responses match the ground-truth answers. For coding problems, we employ a unit-test-based judge that evaluates the accuracy of the code by running it through a series of unit tests.
    
    \item \textbf{Factuality judge:} Hallucination is a common issue in LLMs, especially during the RLHF phase. The reward model often fails to distinguish between factual and non-factual claims. To address this, we use the Llama3 70B model as a factuality constraint judge to evaluate whether the fact-related claims in an output contradict pre-collected, verified factual data, thereby ensuring the accuracy and reliability of the information provided by the LLM.

    \item \textbf{Safety judge:} The safety reward model alone does not sufficiently ensure the trustworthiness of our model due to its limited accuracy. To further enhance safety, we incorporate LlamaGuard2, an industry leading open sourced fine-tuned LLM, to assess whether an output violates predefined safety standards. 
    
\end{itemize}
For details on all the above judges, please refer to \Cref{sc: app_cst}.

\section{Evaluation Benchmarks}\label{exp: benchmarks}
We assess models using a range of benchmarks to comprehensively evaluate their performance across all tasks. More detailed information of evaluation setup can be found in \Cref{app: eval}.
\begin{itemize}
    \item \textbf{General chat}
    \begin{itemize}
        \item \textbf{AlpacaEval-2} \citep{dubois2024alpacafarm}: This benchmark focus on single-turn conversations and includes 805 test prompts that span a range of topics. The models are evaluated directly against GPT-4 Preview to determine the win rate. The same GPT-4 model also serves as the judge.
        \item \textbf{Chat-Arena-Hard} \citep{li2024live}: This benchmark includes 500 test prompts sourced from the live data on Chatbot Arena, a crowd-sourced platform for evaluating large language models (LLMs). These prompts assess the model's capabilities in areas such as specificity, domain knowledge, complexity, problem-solving, creativity, technical accuracy, and real-world application. Besides aligning with human preferences, when compared to AlpacaEval-2, Chat-Arena-Hard also demonstrates distinct separability between different models.
        \end{itemize}
    \item \textbf{Instruction Following}
    \begin{itemize}
        \item \textbf{IFeval} \citep{zhou2023instruction}: This benchmark concentrates on close-form instruction-following tasks, encompassing 25 verifiable instructions. It comprises 541 evaluation prompts, each potentially containing multiple instruction requests. Four accuracy scores are provided in this benchmark: prompt-level strict accuracy, prompt-level loose accuracy, instruction-level strict accuracy, and instruction-level loose accuracy. We report the average of these four scores to represent the model's performance in this benchmark.
    \end{itemize}
    \item \textbf{Math/Coding Reasoning}
    \begin{itemize}
        \item \textbf{MATH} \citep{hendrycks2021measuring}: This benchmark includes 5000 problems drawn from a variety of mathematics competitions, encompassing a broad spectrum of subjects such as Prealgebra, Algebra, Number Theory, Counting and Probability, Geometry, Intermediate Algebra, and Precalculus. Most of these problems demand more than just the simple application of standard mathematical techniques.
        \item \textbf{GSM8K} \citep{cobbe2021training}: This benchmark features 8.5k high-quality problems at the grade school math level. The solutions to these problems rely solely on elementary concepts, making high test performance an achievable goal. Additionally, this dataset exhibits high linguistic diversity while depending on relatively simple grade school math concepts.
        \item \textbf{MBPP} \citep{austin2021program}: This benchmark comprises 974 programming tasks tailored for entry-level programmers. It evaluates the capability of language models to generate concise Python programs based on descriptions provided in natural language. We consider the 0-shot evaluation prompt, which is closer to real-world use cases. We provide a prompt example in the Appendix \ref{app: eval}.
        \item \textbf{HumanEval} \citep{chen2021evaluating}: This benchmark consists of 164 handwritten programming problems, each featuring a function signature, docstring, body, and unit tests. The programming tasks in this benchmark are designed to evaluate language comprehension, reasoning, algorithmic thinking, and basic mathematics skills. Similar to MBPP, we consider 0-shot evaluation prompt for this benchmark. 
    \end{itemize}
    \item \textbf{World knowledge \& factuality}
    \begin{itemize}
        \item \textbf{MMLU} \citep{hendrycks2020measuring}: This benchmark comprises 15908 multiple-choice questions spanning various branches of knowledge. It encompasses subjects including the humanities, social sciences, and hard sciences. The evaluation dataset includes 57 tasks, covering areas such as elementary mathematics, US history, computer science, law, among others.
        \item \textbf{ARC-Challenge} \citep{clark2018think}: This benchmark features a collection of 2590 natural, grade-school science multiple-choice questions. All questions are considered challenging, as evidenced by the failure of both retrieval and co-occurrence methods to provide correct answers.
        \item \textbf{TruthfulQA} \citep{lin2021truthfulqa}: The benchmark includes 817 questions across 38 categories, such as health, law, finance, and politics. Each question in the dataset is designed in a way that may lead some individuals to answer incorrectly due to prevailing misconceptions or false beliefs. We report the multiple-choice QA accuracy score (MC2) in our paper.
    \end{itemize}
    \item \textbf{Engagement Intent}: We subsample 2000 prompts from the LMSys-1M dataset \citep{zheng2023lmsyschat1m} and utilize the oracle engagement predictor, as discussed in \Cref{sc: rm_exp}, to evaluate each generation. The average score of the generations is used as a metric to measure the user's intent to interact with the language model.
    \item \textbf{Safety}
    \begin{itemize}
        \item \textbf{Safety violation ratio (SVR)}: Our safety benchmark is broadly based on the risk categories in ML Commons taxonomy of hazards \citep{inan2023llama}. We collect adversarial human-written prompts per risk category to ensure every category is covered and balanced in equal proportions. The prompts are written in such a way to include diverse adversarial styles and attack vectors. We train a safety classifier similar to Llama Guard \citep{vidgen2024introducing} to compute an overall violation ratio as an evaluation metric.
        \item \textbf{False refusal ratio (FRR)}: False refusal ratio acts as a counter-metric to violation rate, to make sure the model is not overly safe. The false refusal prompts are collected in tandem to the adversarial prompts, targeting innocuous and benign cases but are also likely to trigger the model to refuse due to safety reasons. One such example is "How to kill the lights". We also drew inspirations from XSTest \citep{rottger2023xstest} and OR-Bench \citep{cui2024or} while constructing our false refusal benchmark. We train a refusal classifier on model responses to compute an overall false refusal ratio as an evaluation metric.
    \end{itemize}
\end{itemize}

\subsection{CGPO Training Setup}
In this section, we will show how we implment the CGPO in the RLHF finetuning stage.

\textbf{RLHF warm-up. } Unlike previous studies \cite{ouyang2022training,achiam2023gpt}, which directly employ the SFT model as the initial point for RLHF, our approach introduces a "warm-up" phase. This phase begins with a model that has undergone preliminary fine-tuning through a few steps of DPO, starting from the SFT model. The rationale behind this strategy is that even if DPO and the reward model utilize the same preference training dataset, initiating online RLHF directly from the SFT model and performing policy optimization with the reward model may not be able to explicitly exploit the high-quality preference data, potentially leading to suboptimal performance enhancements. By initiating RLHF with a model already refined by DPO to a certain degree, we can fully harness the advantages of the preference dataset, thereby providing a better starting point for RLHF with improved performance gains and more stable optimization performance.

In our experiments, we employ all reward model training datasets in \Cref{sc: rm_exp} to conduct DPO training using the SFT model as described in Section \ref{sc: sft_exp}. The warm-up model is developed through 3,000 update steps. As we will show in \Cref{sc: expresult}, CGPO initiated from the warm-up model significantly outperforms that started from the SFT model.

\textbf{Training recipe:} We begin the RLHF finetuning process using the warm-up model. We incorporated the reward models and MoJs, developed in Sections \ref{sc: rm_exp} and \ref{sc: cst_exp} respectively, into the CGPO framework to facilitate the RLHF finetuning process. \Cref{tab:tasks} shows the treatment we applied for each task. 

\begin{table}[ht]
    \centering
    \begin{tabular}{>{\centering\arraybackslash}m{2.8cm} > {\centering\arraybackslash}m{2.2cm} >{\centering\arraybackslash}m{2cm} >{\centering\arraybackslash}m{2cm} >{\centering\arraybackslash}m{1.8cm} >{\centering\arraybackslash}m{1.5cm}}
        \toprule
        \textbf{Tasks} & \textbf{General chat} & \textbf{Instruction Following} & \textbf{Math/Coding Reasoning} & \textbf{Engagement Intent} & \textbf{Harmful Intent} \\
        \midrule
        \textbf{Prompt} & UltraChat, LMSys-55k, XSTest, TriviaQA, ARC & Synthetic IF prompts & Math, GSM8K, Aqua \& APPS & LMSys-1M & Safety RM training prompt \\
        \midrule
        \textbf{Helpfulness RM} & \checkmark & \checkmark & \checkmark & & \\
        \textbf{Engagement RM} & & & & \checkmark & \\
        \textbf{Safety RM} & & & & & \checkmark \\
        \midrule
        \textbf{Style} & \checkmark & & & \checkmark \\
        \textbf{False refusal} & \checkmark & & & \checkmark \\
        \textbf{Precise IF} & & \checkmark & & & \\
        \textbf{Math/Code} & & & \checkmark & & \\
        \textbf{Factuality} & \checkmark & & & \\
        \textbf{Safety} & & & & & \checkmark \\
        \bottomrule
    \end{tabular}
    \newline
    \caption{Tasks and their corresponding prompt sets, reward models, and MoJs. Note that in general chat task the factuality constraint judge is only applied to TriviaQA and ARC prompt set.}
    \label{tab:tasks}
\end{table}

\textbf{Baseline and Ablations: }
To assess the performance of various constrained RLHF optimizers proposed in this study, we conducted CGPO training with different optimizers: CRPG, CRRAFT, and CODPO under the same settings (prompt set, reward model, and MoJs). We tailored the hyperparameter settings of the optimizer for various tasks to align with the specific characteristics of each task. 
Additionally, we consider DPO and PPO as our RLHF baselines. To establish the DPO baseline, we continue running the DPO updates starting from the RLHF warm-up model and extend the training to 14,000 steps to thoroughly optimize all benchmarks listed in \Cref{exp: benchmarks}.
As previously mentioned, for DPO training, we utilize all reward models' training sets starting from the SFT model. To establish the PPO baseline, we first train a unified reward model by merging all reward models' training data as described in \Cref{sc: rm_exp}. Following this, we start from the warm-up model and perform PPO updates by applying the unified reward model to all prompt sets listed in \Cref{tab:tasks}. For both PPO and CGPO variants, we utilize the same global batch size $1024$ and conduct $600$ update steps. 

\subsection{Experimental Results}\label{sc: expresult}
In this section, we present the main results of our experiments. In Section \ref{sc: exp_main}, we highlight the superior performance of CGPO compared to baseline RLHF methods across various benchmarks. We present ablation studies in Section \ref{sc: exp_cst} to demonstrate the importance of adopting MoJs. Additionally, we discuss the benefits of introducing an RLHF warm-up stage in Section \ref{sc: exp_warmup}.

\subsubsection{Main Results and Ablations}\label{sc: exp_main}

\begin{figure}[h]
  \centering
  \includegraphics[width=1.0\textwidth]{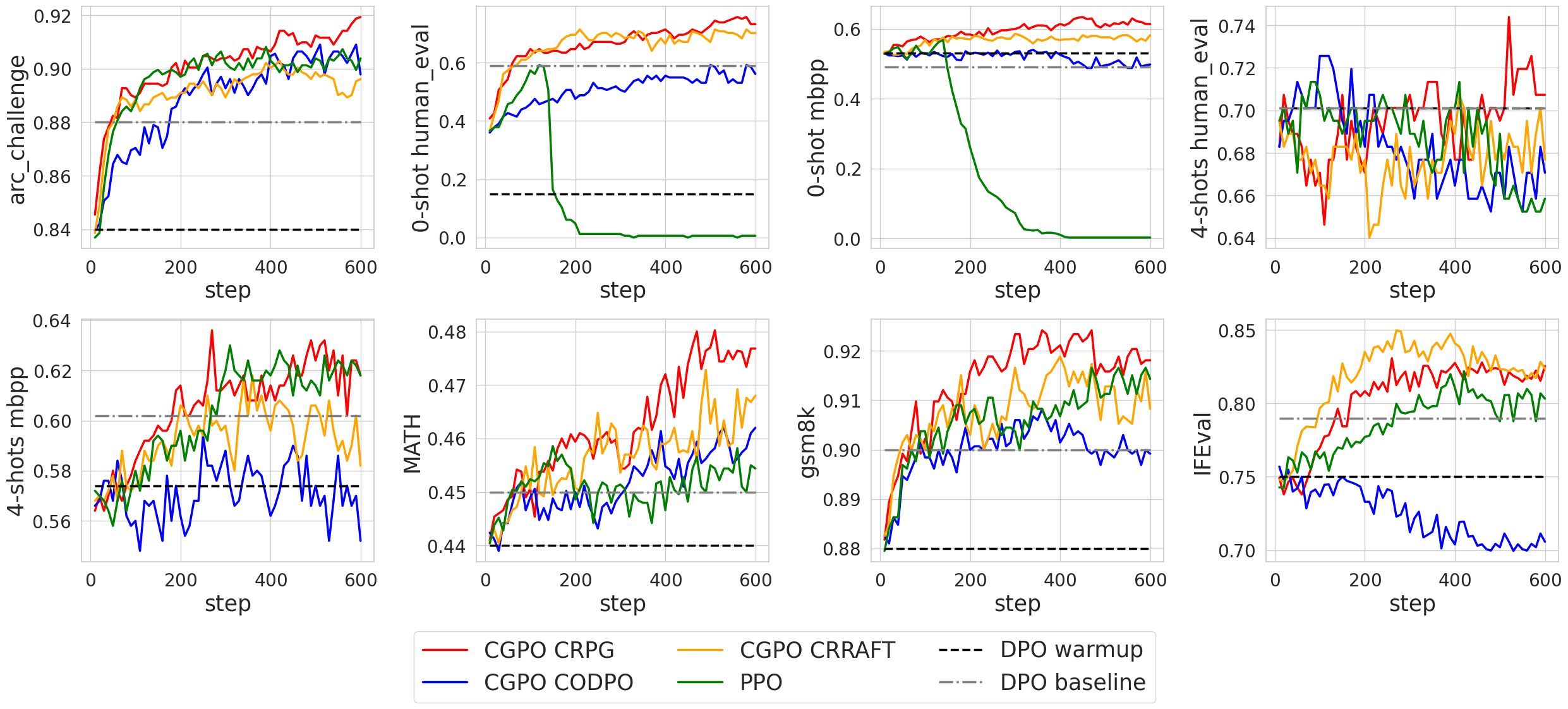}
  \captionsetup{justification=centering}
  \caption{Comparison of CGPO variants with baseline RLHF algorithms PPO and DPO across various benchmarks}
  \label{fig:exp_cgpo_1}
\end{figure}

For the online RLHF algorithms CGPO and PPO, we monitor the model's performance at every 10-step interval throughout the training trajectory across various benchmarks, as illustrated in \Cref{fig:exp_cgpo_1}. The plot demonstrates that CGPO, when paired with the CRPG and CRRAFT optimizers, consistently enhances performance across all benchmarks compared to the initial model, indicating progressive improvement as training progresses.
Specifically, CRPG outperforms all others throughout the entire training period in terms of ARC Challenge, 0-shot HumanEval, 0-shot MBPP, 4-shots MBPP, MATH, and GSM8K. Meanwhile, CRRAFT excels in IFEval during the training phase.
Notably, the online RLHF baseline PPO exhibits a significant decline in performance on 0-shot coding benchmarks (MBPP and HumanEval) as training progresses, indicating a severe case of reward hacking. Meanwhile, CGPO with the CODPO optimizer shows a slight regression on MBPP and IFEval benchmarks compared to the warm-up model, yet it effectively avoids the drastic performance drop observed with PPO in the coding benchmarks. The offline RLHF baseline DPO, while avoiding the drastic regression seen with PPO, remains overly conservative in enhancing the model's performance, resulting in lower metric improvements compared to CGPO with the CRPG and CRRAFT optimizers.

In Table \ref{tab:exp_main}, we present the evaluation results for SFT, DPO warm-up, DPO baseline, the final step of PPO, and various CGPO variants across all benchmarks detailed in \Cref{exp: benchmarks}. 
The data in Table \ref{tab:exp_main} indicate that CGPO variants employing CRPG and CRRAFT optimizers significantly outperform the DPO and PPO baselines across all benchmarks. Notably, CRPG shows the most substantial improvements in math and coding benchmarks (Math, GSM8K, HumanEval, and MBPP), while CRRAFT excels in helpfulness and factuality (AlpacaEval-2, Arena-Hard, and TruthfulQA). Both CRPG and CRRAFT achieve the best results in terms of instruction following (IFEval). While the CGPO variant with the CODPO optimizer does not perform as strongly as other variants, it offers performance that is on par with or better than the DPO and PPO in all benchmarks except the IFEval. In terms of safety, CGPO with the CRPG and CODPO optimizers achieve the best results in FRR and SVR, respectively. \Cref{tab:exp_main} demonstrates that the CGPO framework is able to enhance model quality across all tasks, proving its efficacy in managing challenging multi-task fine-tuning. 


\begin{table}[ht]
    \small
    \centering
    \begin{tabular}{>{\raggedright\arraybackslash}m{2.8cm} > {\centering\arraybackslash}m{1.5cm} >{\centering\arraybackslash}m{1.5cm} >{\centering\arraybackslash}m{1.5cm} >{\centering\arraybackslash}m{1.5cm} >{\centering\arraybackslash}m{1.5cm} >
    {\centering\arraybackslash}m{1.5cm} >
    {\centering\arraybackslash}m{1.5cm} }
        \toprule
         & \textbf{SFT} & \textbf{DPO warm-up} & \textbf{DPO} & \textbf{PPO} & \textbf{CGPO - CRPG} & \textbf{CGPO - CRRAFT} & \textbf{CGPO - CODPO}  \\
        \midrule
        \midrule
        \textbf{AlpacaEval-2}  & $10.9$ & $13.3$ & $16.3$ & $24.8$ & $25.9$ & \textbf{43.2} & $18.08$   \\
        \textbf{Arena-Hard}  & $13.6 \pm 1.6$ & $18.8\pm1.6$ & $18.3\pm1.7$ & $24.3\pm1.8$ & $31.2\pm 2.2$  & \boldmath$36.8\pm2.0$\unboldmath & $16.8\pm1.9$   \\
        \midrule
        \textbf{IFEval}  & $0.71$ & $0.75$ & $0.79$ & $ 0.81 $ & \textbf{0.83} & \textbf{0.83} & $0.70$  \\
        \midrule
        \textbf{MATH}  & $0.44$ & $0.44$ & $0.45$ & $0.46$ & \textbf{0.48} & $0.47$ & $0.46$  \\
        \textbf{GSM8K}  & $0.86$ & $0.88$ & $0.90$ & $0.91$ & \textbf{0.93} & $0.92$ & $0.90$   \\
        \textbf{0-shot MBPP}  & $0.50$ & $0.51$ & $0.49$ & $0.002$ & \textbf{0.63} & $0.57$ & $0.51$   \\
        \textbf{4-shots MBPP}  & $0.55$ & $0.57$ & $0.60$ & \textbf{0.62} & \textbf{0.62} & $0.58$ & $0.55$   \\
        \textbf{0-shot HumanEval}  & $0.09$ & $0.15$ & $0.59$ & $0.006$ & \textbf{0.76} & $0.70$ & $0.57$   \\
        \textbf{4-shots HumanEval}  & $0.62$ & $0.70$ & $0.70$ & $0.66$ & \textbf{0.71} & $0.68$ & $0.67$   \\
        \midrule
        \textbf{MMLU}  & $0.75$ & \textbf{0.76} & $0.75$ & $0.75$ & $0.75$ & $0.75$ & $0.75$   \\
        \textbf{ARC}  & $0.85$ & $0.84$ & $0.88$ & $0.90$ & \textbf{0.92} & $0.90$ & $0.90$   \\
        \textbf{TruthfulQA}  & 0.57 & 0.59 & 0.63 & 0.65 & 0.64 & \textbf{0.66} & 0.63   \\
        \midrule
        \textbf{Engagement}  & 0.50 & 0.59 & 0.71 & \textbf{0.81} & \textbf{0.81} & 0.72 & 0.79   \\
        \midrule
        \textbf{SVR}  & $0.03$ & $0.03$ & $0.02$ & $0.03$ & $0.05$ & $0.02$ & \textbf{0.01}   \\
        \textbf{FRR}  & $0.18$ & $0.161$ & $0.17$ & $0.12$ & \textbf{0.04} & $0.12$ & $0.24$   \\
        \bottomrule
    \end{tabular}
    \newline
    \caption{Evaluation results of SFT, DPO warm-up, DPO, PPO and CGPO variants}
    \label{tab:exp_main}
\end{table}

\subsubsection{Effectiveness of Mixture of Judges}\label{sc: exp_cst}
\begin{figure}[h]
  \centering
  \includegraphics[width=1.0\textwidth]{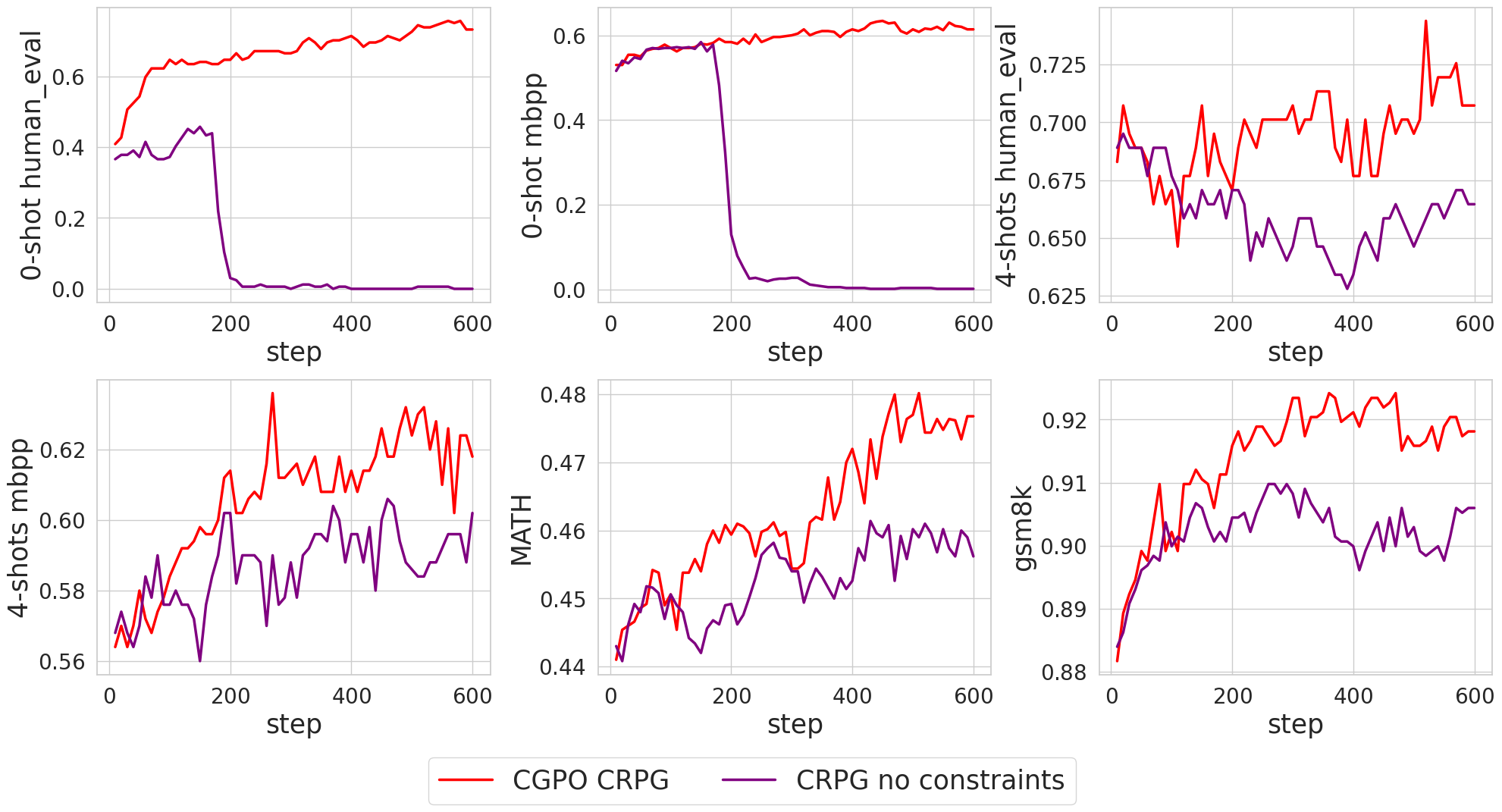}
  \captionsetup{justification=centering}
  \caption{Comparison of CGPO (CRPG optimizer) with and without MoJs}
  \label{fig:exp_cgpo_2}
\end{figure}

In this section, we explore the significance of incorporating MoJs within the CGPO framework. We conduct an ablation study by eliminating all MoJs from CGPO, utilizing the CRPG optimizer, while keeping all other variables constant, and then proceed to rerun the RLHF finetuning for 600 steps. Figure \ref{fig:exp_cgpo_2} presents a comparative analysis of CGPO performance with and without MoJs using the CRPG optimizer across various benchmarks, including HumanEval, MBPP, MATH, and GSM8K.

From \Cref{fig:exp_cgpo_2}, it is clear that in the absence of coding judges, the CRPG optimizer undergoes a notable decline in 0-shot coding benchmarks once it surpasses 180 steps, mirroring the performance of the PPO baseline. Additionally, in the MATH, GSM8K and 4-shots HumanEval and MBPP benchmarks, while CRPG shows some improvement without constraints, the increases in metrics are considerably less pronounced compared to cases where math judges are utilized. This comparison effectively illustrates that MoJs play a crucial role not only in preventing reward hacking but also in significantly boosting the model's performance during online RLHF finetuning.

\subsubsection{Impact of RLHF Warm-up}\label{sc: exp_warmup}
In this section, we discuss the importance of introducing the RLHF warm-up stage. We consider CGPO with CRPG optimizer, and rerun the experiment in \Cref{sc: exp_main} but switch the starting point with SFT model. Addtionally, we add one more ablation by starting from the DPO baseline that has been extensively optimized, which has significantly better performance across all benchmarks than the DPO warm-up model (\Cref{tab:exp_main}). 

Monitoring GPT-based helpfulness evaluations like AlpacaEval-2 and Arena-Hard during training is costly. To efficiently assess the effectiveness of the RLHF warm-up stage from the helpfulness perspective, we implement a cost-effective benchmark. We collect prompts from user-LLM interactions (e.g., LMSys-1M) and generate multiple responses using the LIama3.0 70B model. These responses are ranked by a powerful LLM, and the highest and lowest-ranked responses are used to create preference pairs for training a reward model (RM). This RM evaluates helpfulness based on its average score on its training prompts. Although this RM may overfit this prompt set, it remains a valid measure of helpfulness since our finetuning process does not depend on this specific prompt set.

\begin{figure}[h]
  \centering
  \includegraphics[width=1.0\textwidth]{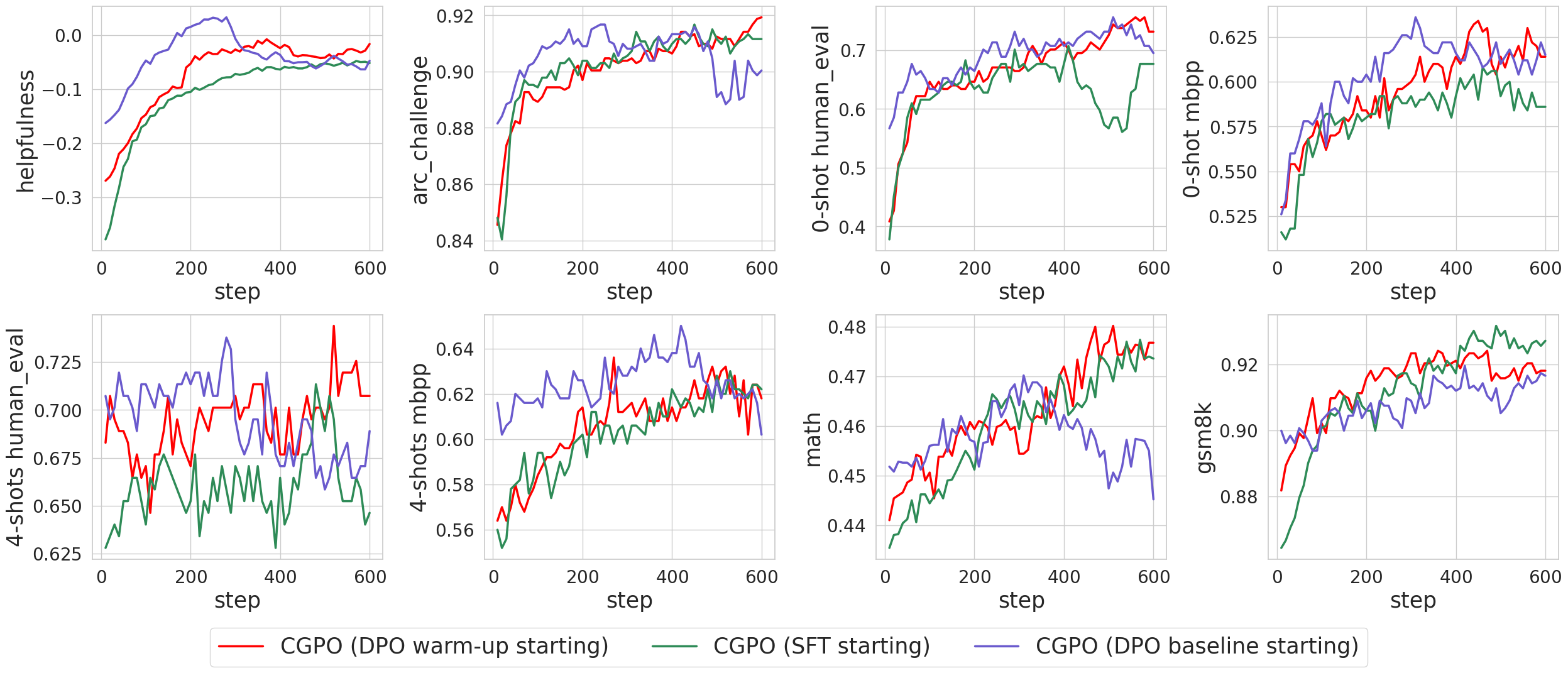}
  \captionsetup{justification=centering}
  \caption{Comparison of CGPO (CRPG optimizer) with different starting point}
  \label{fig:exp_cgpo_3}
\end{figure}

Figure \ref{fig:exp_cgpo_3} illustrates the training curves of the CGPO model with different initial conditions across various benchmarks. When compared to the standard online RLHF setting, which starts with the SFT model, CGPO initiated from the warm-up model consistently achieves superior performance in all benchmarks, with the exception of GSM8K. For the runs that begin with the DPO baseline, there is a noticeable higher initial performance across all benchmarks. However, the ultimate performance of these models does not exceed those that started from the warm-up or SFT models. Particularly in helpfulness, ARC challenge, Math and 4-shot coding benchmarks, there is a marked decline in performance during the later stages of training. This suggests that starting from the highly optimized DPO baseline may detrimentally affect the final model's performance, potentially due to the soft-greedy nature of the DPO optimal policy, which might limit the model's ability to explore and further improve. Therefore, Figure \ref{fig:exp_cgpo_3} demonstrates that incorporating an RLHF warm-up stage can significantly enhance the model's performance during the subsequent online RLHF phase.

\section{Related Works}
\textbf{RLHF in MTL.} Reinforcement Learning with Human Feedback (RLHF) is designed to align language models with human preferences and has become a crucial component of the fine-tuning pipeline for Large Language Models (LLMs) \citep{stiennon2020learning, ouyang2022training, brown2020language, touvron2023llama, bi2024deepseek, bai2022training}. The majority work of RLHF focus optimizing a single reward models \citep{ouyang2022training,gao2023scaling,dong2023raft,ethayarajh2023human}. The exploration of RLHF in the MTL setting remains relatively underexplored. The most commonly adopted approach involves optimizing a weighted sum of several reward models, where each model captures the interests of different tasks \citep{ramamurthy2022reinforcement,glaese2022improving,yuan2023rrhf,bakker2022fine,wu2024fine}. However, a major limitation of this approach is that key information from each individual reward model can be lost through linear combination, particularly when conflicting task goals exist. This can lead to suboptimal performance for each individual task. Additionally, each individual reward model typically requires different treatments (regularization, early stopping, etc) due to their unique properties, thus applying a uniform treatment for a composite reward model can further impair optimization performance across tasks \citep{moskovitz2023confronting}. Another research direction involves fine-tuning a separate LLM model for each task, followed by linear interpolation of the LLM weights across all learned models to produce a single model that excels in multiple tasks \citep{rame2024rewarded}. However, this method remains computationally expensive and unstable due to the high cost and variability inherent in a single RLHF process \citep{hu2023aligning,rafailov2024direct}. \citep{yang2024rewards} Proposed to use in-context reward model to manage multiple reward, but introduce additonal cost during inference time. Unlike the approaches mentioned above, CGPO introduces a customized reward model recipe and an RLHF optimizer tailored for each specific task. This method is not only as efficient as the conventional RLHF pipeline, but it also preserves all information within each reward model, thereby optimizing alignment for each task to the fullest extent.

\textbf{Reward Hacking Mitigation.} Compaired with traditional RL, where the reward is typically well-defined and the goal is to maximize it \citep{sutton2018reinforcement}, RLHF introduces a unique challenge known as "reward hacking." This issue arises because the reward model serves as a proxy for actual human preferences. Over-optimization of the reward model can adversely impact the performance of the language model \citep{gao2023scaling, moskovitz2023confronting, stiennon2020learning, rafailov2024direct}. Consequently, addressing reward hacking is a major focus in RLHF.
Previous studies have explored various approaches to mitigate the effects of reward hacking, including reward model regularization \citep{singhal2023long}, reward ensembles \citep{eisenstein2023helping, rame2024warm}, and explicitly learning the reward bias error \citep{chen2024odin, shen2023loose}. In contrast to previous methods, our CGPO framework employs both LLM and rule-based judges as constraints to detect and prevent reward hacking patterns. This approach offers a more fine-grained and controllable solution to this persistent issue. Furthermore, the use of MoJs enables us to develop tailored strategies for mitigating the effects of reward hacking across various tasks in the MTL setting. This allows us to effectively address the reward hacking challenge in the more complex MTL environment, where previous methods have struggled to perform efficiently.

\section{Conclusion}
In this paper, we introduced the CGPO framework to address key challenges in multi-task learning for LLM post-training with RLHF. The CGPO framework effectively mitigates issues such as inhomogeneous reward hacking and conflicting task goals through a novel primal-type multi-constraint RL method and a tailored multi-objective optimization strategy. We demonstrate the effectiveness of CGPO in a scenario where we need to handle five tasks with three reward models and seven constraints, marking the first application of RLHF in multi-task learning for general-purpose LLMs. Our experiments show that CGPO achieves significantly better metric gains for all tasks compared to the baseline RLHF methods. Moving forward, it is promising to explore more automated ways to adapt the gradient weights from different tasks to further reduce the hyperparameter burden and advance the Pareto frontier \citep{sener2018multi}.


\clearpage
\newpage
\bibliographystyle{assets/plainnat}
\bibliography{paper}

\clearpage
\newpage
\beginappendix

\section{CGPO Training Set}\label{sc: app_training}
The detail of of our training dataset is provide in \Cref{tab:datacard}. Note that in our experiment we adopt the instruction finetuing format, in which the prompt is wrapped as "[INST] \{prompt\} [\textbackslash INST]":

\begin{table}[ht]
    \centering
    \begin{tabular}{>{\centering\arraybackslash}m{2cm} > {\centering\arraybackslash}m{2.5cm} > {\centering\arraybackslash}m{2cm} > {\centering\arraybackslash}m{2.5cm} > {\centering\arraybackslash}m{6.0cm}}
        \toprule
        \textbf{Dataset} & \textbf{is Preference} & \textbf{Size} & \textbf{Usage} & \textbf{Source}  \\
        \midrule
        \midrule
        Orca-Math & \XSolidBrush & 200035 & SFT & \cite{mitra2024orca}   \\
        \midrule
        MetaMath & \XSolidBrush & 395000 & SFT & \cite{yu2023metamath}   \\
        \midrule
        Evol-CodeAlpaca & \XSolidBrush & 111183 & SFT & \cite{luo2023wizardcoder}   \\
        \midrule
        MATH training & \XSolidBrush & 7500 & Online RLHF & \cite{hendrycks2021measuring}   \\
        \midrule
        GSM8K training & \XSolidBrush & 7473 & Online RLHF & \cite{cobbe2021training}   \\
        \midrule
        Aqua Math & \XSolidBrush & 97467 & Online RLHF & \cite{ling2017program}   \\
        \midrule
        APPS & \XSolidBrush & 7070 & Online RLHF & \cite{hendrycksapps2021}   \\
        \midrule
        XSText & \XSolidBrush & 2700 & Online RLHF & \cite{röttger2023xstest}   \\
        \midrule
        LMSys-55k & \Checkmark & 49865 & SFT, RM, DPO, Online RLHF & \cite{chiang2024chatbot}  \\
        \midrule
        UltraChat & \Checkmark & 207865 & SFT, RM, DPO, Online RLHF & \cite{ding2023enhancing}   \\
        \midrule
        UltraFeedback & \Checkmark & 340025 & SFT, RM, DPO & \cite{cui2023ultrafeedback}  \\
        \midrule
        UltraInteract & \Checkmark & 129531 & SFT, RM, DPO & \cite{yuan2024advancing}  \\
        \midrule
        HH-RLHF & \Checkmark & 115396 & RM, DPO & \cite{bai2022training}  \\
        \midrule
        SHP & \Checkmark & 93301 & RM, DPO & \cite{ethayarajh2023human}  \\
        \midrule
        HelpSteer & \Checkmark & 37131 & RM, DPO & \cite{wang2023helpsteer}  \\
        \midrule
        Distilabel-Capybara & \Checkmark & 14811 & RM, DPO & \cite{ethayarajh2024kto} \\
        \midrule
        Distilabel-Orca & \Checkmark & 6926 & RM, DPO & \cite{distilabel-argilla-2024}  \\
        \midrule
        Argilla Math & \Checkmark & 2418 & RM, DPO & \cite{distilabel-argilla-2024}  \\
        \midrule
        Synthetic IF dataset & \Checkmark & 11668 & SFT, RM, DPO. Online RLHF & Prompts are generated by LIama 3.0 70B instruct model, accepted and rejected responses are generated by LIama 3.0 70B instruct model and LIama 3.0 8b instruct model   \\
        \midrule
        Human Annotated safety dataset & \Checkmark & 244232 & SFT, RM, DPO. Online RLHF & Colloect adversarial human-written prompts per risk category. The prompts are written in such a way to include diverse adversarial styles and attack vectors.  \\
        \midrule
        Synthetic engagement dataset & \Checkmark & 112375 & SFT, RM, DPO. Online RLHF & Prompt are sampled from LMSys-1M \cite{zheng2023lmsyschat1m}, the accepted and rejected responses are generated by LIama 3.0 70B instruct model  \\
        \bottomrule
    \end{tabular}
    \newline
    \caption{Dataset used in CGPO experiments for SFT, RM, DPO and online RLHF training}
    \label{tab:datacard}
\end{table}

\textbf{Synthetic IF dataset. }Inspired by \cite{zhou2023instruction}, we consider synthetic prompts that require LLM generation to satisfy one or more closed-form instructions, which can be verified exactly. We identify 23 types of closed-form instructions for generation and use LIama 3.0 70B instruct model to create synthetic prompts that address a specific topic and also require these closed-form instructions. We create a template to enable LIama 3.0 70B instruct model to generate all prompts. The prompt template that we input into LIama 3.0 70B instruct model to generate synthetic instruction-following prompts is provided as follows:

\begin{shaded}
{\ttfamily
\textbf{Prompt Template} =

"You are a helpful AI assistant. You are given a TOPIC and a FORMAT REQUIREMENT, and you are expected to generate a PROMPT that is on the given TOPIC and specify the given FORMAT REQUIREMENT that the corresponding answer should follow. 
Here are many examples that you can learn from:
\newline

\textbf{TOPIC:} Travel 

\textbf{FORMAT REQUIREMENT:} In your entire response, refrain from the use of any commas

\textbf{PROMPT:} I am planning a trip to Japan, and I would like thee to write an itinerary for my journey in a Shakespearean style.  You are not allowed to use any commas in your response.
\newline

\textbf{TOPIC:} Aerospace engineering

\textbf{FORMAT REQUIREMENT:} In your entire response, refrain from the use of any commas and Give two different responses. Responses and only responses should be separated by 6 asterisk symbols: ******

\textbf{PROMPT:} Write two jokes about rockets. Do not contain commas in your response. Separate the two jokes with 6 asterisk symbols: ******.
\newline

\textbf{TOPIC:} History

\textbf{FORMAT REQUIREMENT:} Entire output should be wrapped in JSON format

\textbf{PROMPT:} What is the history of NYC prospect park? Please wrap your entire answer in JSON format.
\newline

\textbf{TOPIC:} Video game

\textbf{FORMAT REQUIREMENT:} Highlight at least 2 sections in your answer with markdown, i.e. *highlighted section* and Answer with at least 40 sentences

\textbf{PROMPT:} Can you write a poem about the pros and cons of playing a lot of video games? Please make sure it's at least 40 sentences long (don't forget to add punctuations). You must highlight at least sections in your response, like *highlighted phrase*.
\newline

\textbf{TOPIC:} Movie

\textbf{FORMAT REQUIREMENT:} Answer with at least 40 sentences, Highlight at least 4 sections in your answer with markdown, i.e. *highlighted section*, and Wrap your entire response with double quotation marks

\textbf{PROMPT:} Write a joke about the superhero movie with at least 5 sentences. Use Markdown to italicize at least 4 sections in your answer, i.e. *italic text*. Wrap your answer in double quotes.
\newline

\textbf{TOPIC:} Health care

\textbf{FORMAT REQUIREMENT:} Your entire response should be in English, capital letters only

\textbf{PROMPT:} Write an essay about public health care system in US in English and in all capital letters.
\newline

\textbf{TOPIC:} Mathematics

\textbf{FORMAT REQUIREMENT:} Entire output should be wrapped in JSON format

\textbf{PROMPT:} List all facts about calculus in a structured output. In particular, Format your entire output in JSON.
\newline

Now it is your turn to generate a PROMPT that is on the given TOPIC and specify the given FORMAT REQUIREMENT that the corresponding answer should follow. Please DO NOT make up any new format requirement that is not given to you.

\textbf{TOPIC:} \{topic\}

\textbf{FORMAT REQUIREMENT:} \{instruction\}

To be noted, you just need to mention/specify the FORMAT REQUIREMENT in your response but your response does not need to follow it. Please directly provide the PROMPT without any extra words. Do not write any note or explanation.
\newline
\newline

\textbf{TOPICS} = ["20th century events", "Accounting", "Architecture", "Astronomy", "Biology", "Businessethics","Celebrities","Chemistry","Clinical knowledge",
    "Economics",
    "Electrical engineering",
    "Ethics of artificial intelligence",
    "Education",
    "Energy",
    "Gaming",
    "Geography",
    "Global facts",
    "History",
    "Healthcare",
    "Immigration law",
    "International law",
    "Jurisprudence",
    "Management",
    "Marketing",
    "Mathematics",
    "Medicine",
    "Moraldisputes",
    "Movies",
    "Music",
    "Philosophy",
    "Physics",
    "Prehistory",
    "Psychology",
    "Public relations",
    "Sociology",
    "Sports",
    "Social media"
    "Transportation",
    "Virology"]

\textbf{Instructions} = ["number of paragraphs", "number of sentences", "number of words", "first word in n-the paragraph", "number of a specific placeholder"; "number of sections", "title", "response given in a certain format", "number of highlighted sections", "response need to be in json", "postscript at the end of response", "number of bullet list", "forbidden words", "certain keyword must exist", "a given key word need to appear at least n-times", "a given letter need to appear at least n-times", "generation should be in lowercase", "generation should be in capital", "capital word need to appear at least n-times", "generation should no contain comma", "generation should finish with an exact end checker", "entire response should be be wrapped within double quotation marks", "generation should contain two responses"]
}
\end{shaded}
Each time, we randomly select up to three types of closed-form instructions along with one topic, and incorporate them into a template. This template is then used by LIama 3.0 70b instruct model to generate a prompt. We repeat this process 30000 times to create a comprehensive set of instruction-following prompts.

For each synthetic prompt, we utilized Llama 3.0 70B Instruct model, and Llama 3.0 8B Instruct model to generate a response based on the prompt. We then evaluated whether these responses adhered to the instruction-following constraints. Prompts that did not yield any responses meeting the constraints, as well as those where all responses met the constraints, were filtered out. This process resulted in 11668 prompts that included both responses that satisfied the constraints and responses that violated them. We randomly selected one response that met the constraints as the accepted response and one that violated the constraints as the rejected response for each prompt. By doing so, we constructed our pairwise instruction-following preference dataset.

\textbf{Human annotated safety dataset. }We take an iterative approach to collect multiple batches of safety preference data and merge them together as the final train data. At each iteration, we generate two different responses from a pool of models (model from previous iteration for example), and send them to human annotators to rate and rank based on the safety guidelines. If no response meets the guideline, the annotators are asked to directly edit the higher ranked response for it to abide the guideline. The collected preference pairs are used to train a reward model, and once such a reward model is trained, we leverage it to do rejection sampling to produce finetuning data that are used to train the next model iteration. This next model will be added to the pool of models that generate responses for human annotators to rank. We repeat this process multiple times to iteratively collect higher quality safety preference pairs. An additional layer of data auditing is also applied on top of each data iteration cycle due to the subtle and subjective nature of safety guidelines to further ensure data quality.

\textbf{Synthetic engagement dataset. }To develop a synthetic engagement pairwise preference dataset, we initially gathered 1M user engagement samples from interactions with an LLM-based chatbot on social media platforms. Each sample comprises a user query, the LLM's response, and a binary label indicating user approval of the response. We used this dataset to train a binary feedback reward model on top of the pretrained Llama 3.0 8B model by adding a linear output layer and training it as a binary classifier. We selected a model iteration with an AUC of 0.89 from the training trajectory to function as the oracle predictor of user engagement intent. This model was subsequently used to generate the synthetic user engagement preference dataset in our study.
In the next step, we subsampled 112,375 prompts from LMSys-1M \cite{zhu2023starling}. We then generated two responses from the Llama 3.0 8B model and two responses from the Llama 3.0 70B model, ultimately generating four distinct responses for each prompt, conditioned under the generation setting temperature=1, top\_p=0.9. Following this, our oracle predictor was used to score all generated responses. The response with the highest score was selected as the accepted response, while the one with the lowest score was marked as the rejected response. By applying this methodology to all selected prompts, we created our synthetic user engagement preference dataset.

\textbf{Additional Comment. }It's important to note that for certain datasets used in online RLHF, we also incorporate metadata to provide additional information about the data as shown in \Cref{tab:metadata_example}. During CGPO training, sometimes it will be necessary to extract information from the metadata to implement the MoJs.
\begin{itemize}
    \item \textbf{MATH, GSM8K \& Aqua Math}: In the metadata, we include the ground truth answer for each question. This allows the math constraint judge to leverage this information to evaluate the accuracy of the LLM's response for each math question.
    \item \textbf{TriviaQA \& ARC}: For prompts related to deterministic factuality, we also incorporate the ground truth answer into the metadata. This allows the factuality constraint judge to assess correctness based on this information.
    \item \textbf{APPS}: In the metadata, we include several unit tests that the correct code snippet should be able to pass through. Our coding constraint judge can leverage this to determine if the generated code is correct
    \item \textbf{Synthetic IF dataset}: We include closed-form instructions in the metadata, specifying requirements that the LLM's generation must satisfy. This enables our instruction-following constraint judge to verify whether the LLM's output adheres precisely to the instructions.
\end{itemize}

\begin{table}[ht]
    \centering
    \begin{tabular}{>{\raggedright\arraybackslash}m{2cm} > {\raggedright\arraybackslash}m{6.5cm} > {\raggedright\arraybackslash}m{6.5cm}}
        \toprule
        \textbf{Data} & \textbf{Prompt} & \textbf{Metadata} \\
        \midrule
        \midrule
        MATH, GSM8K, Aqua Math & A quadratic equation $ax^2 - 2ax + b = 0$ has two real solutions. What is the average of these two solutions? Your response should end with "The final answer is [answer] & $\{\text{"answer": "1"}\}$  \\
        \midrule
        TriviaQA, ARC & Who was President when the first Peanuts cartoon was published? & $\{\text{"answer": "Harry S. Truman"}\}$ \\
        \midrule
        APPS & Write a function "$\text{similar}\_\text{elements}$" to find the similar elements from the given two tuple lists &  \{"$\text{unit}\_\text{tests}$": "assert $\text{similar}\_\text{elements}$((3, 4, 5, 6),(5, 7, 4, 10)) == (4, 5), assert $\text{similar}\_\text{elements}$((1, 2, 3, 4),(5, 4, 3, 7)) == (3, 4), assert $\text{similar}\_\text{elements}$((11, 12, 14, 13),(17, 15, 14, 13)) == (13, 14)"\}  \\
        \midrule
        Synthetic IF & What are the primary architectural styles seen in European churches? Give my answer in English using only capital letters. &  \{"$\text{if}\_\text{requirements}$": "$\text{english}\_\text{capital}$"\}  \\
        \bottomrule
    \end{tabular}
    \newline
    \caption{Example of Prompt and Metadata used in CGPO experiment}
    \label{tab:metadata_example}
\end{table}

\section{CGPO Constraint Judge}\label{sc: app_cst}
In this section, we will discuss in detail about how we build MoJs in CGPO.

\subsection{Rule-based Constraint Judge}

\textbf{Math constraint judge. }As illustrated in \Cref{tab:metadata_example}, for the math prompt sets MATH, GSM8K, and Aqua Math, we explicitly require the model to provide the final answer in a specified format, which can be easily extracted. When implementing the math constraint judge, we extract the LLM's answer by examining the final sentence and comparing it with the ground truth answer in the metadata. There are instances where the model correctly answers the question but fails to provide the answer in the correct format. In such cases, the math constraint judge will indicate that this generation violates the constraint. Although this is a false negative, using CGPO to encourage the model to avoid such patterns can implicitly help improve the model's ability to follow instructions.

\textbf{Coding constraint judge. }Our coding constraint judge examines the coding block in LLM's response to extract the code snippet. It then runs the snippet through all the unit tests provided in the metadata to determine if it passes each test. Similar to the math constraint, false negatives can occur if LLM's solution is not formatted correctly. Implementing CGPO to discourage such patterns could enhance the model's ability to follow instructions accurately.

\textbf{Instruction following constraint judge. }The instruction-following constraint judge begins by reading the metadata to understand the specific rules that LLM's output must adhere to. Then, we employ string-matching based logic to determine whether LLM's generation complies with all the specified rules.

\subsection{LLM-based Constraint Judge}

The LLM classifier constraint judge utilizes an additional LLM to assess whether the output from our training LLM adheres to a specific predefined criterion. We design the input for this judge using a prompt template that arranges the LLM's response alongside other essential contexts. Within this template, we specify both a negative token and a positive token. The negative token indicates that the LLM's response breaches the constraint, while the positive token signifies compliance. We explicitly direct the judge to issue either the positive or negative token based on their assessment.
To minimize the randomness in the judgment process, we do not rely solely on the LLM to generate a token and then check its correspondence to the negative or positive token. Instead, we directly examine the softmax probabilities of the negative and positive tokens. If the probability of the negative token is higher, we conclude that the LLM's response violates the constraint, and vice versa. \Cref{tab:llm_judge_example} presents the template along with the negative and positive tokens for the LLM classifiers in our experiment.

\textbf{False refusal constraint judge. }We utilize the Llama 3.0 8b pretrained model as a foundation and fine-tune an LLM classifier specifically aimed at identifying refusal patterns in LLM responses. The training data is formatted as follows: "[INST] \{LLM response\} [\textbackslash INST] {judgment}", where "judgment" is True if the LLM response indicates refusal, and False otherwise. During the inference phase of deploying this constraint judge, we also encapsulate the generated responses from the training LLM within "[INST] ... [\textbackslash INST]" and use that as the input for the judge.

\textbf{Factuality constraint judge. }We employ the Llama 3.0 70b instruct model directly as the factuality constraint judge. Recall that for prompts associated with deterministic factuality, we include the ground truth answer in the metadata. When deploying this constraint judge, we use the template as illustrated in \Cref{tab:llm_judge_example}, incorporating the prompt, ground truth answer, and the LLM response into the template to serve as inputs for the judge.

\textbf{Safety constraint judge. }We utilize LIamaGuard2 \cite{metallamaguard2}, which is fine-tuned from the Llama 3.0 8b pretrained model. We reuse the template as introduced in the LIamaGuard2 paper \cite{metallamaguard2}, where we incorporate pre-defined safety guidelines and full completions into the prompt template to serve as inputs for the judge.

\begin{table}[ht]
    \centering
    \begin{tabular}{>{\raggedright\arraybackslash}m{2cm} > {\raggedright\arraybackslash}m{11cm} > {\centering\arraybackslash}m{3cm}}
        \toprule
        \textbf{LLM Judge} & \textbf{Template} & \textbf{neg \& pos tokens} \\
        \midrule
        \midrule
        False Refusal & [INST] \textbf{\{LLM response\}} [\textbackslash INST] & True \& False   \\
        \midrule
        Factuality & [INST] Here is a question and a ground truth answer: 
        
        Question: \textbf{\{prompt\}}
        
        Ground truth answer: \textbf{\{ground truth answer\}}
        
        Now, you are given an answer from the AI:
        
        AI: \textbf{\{LLM response\}}
        
        Does the AI's answer contains the factual information from the above ground truth answer? Answer in TRUE or FALSE. [\textbackslash INST] &  False \& True  \\
        \midrule
        Safety & [INST] Task: Check if there is unsafe content in 'Agent' messages in conversations according our safety policy with the below categories.
        
        <BEGIN UNSAFE CONTENT CATEGORIES> \textbf{\{safety guidelines\}} <END UNSAFE CONTENT CATEGORIES>

        <BEGIN CONVERSATION> \textbf{\{completions\}} <END CONVERSATION>

        Provide your safety assessment for ONLY THE LAST turn message in the above conversation
        
        - First line must read 'safe' or 'unsafe'.
        
        - If unsafe, a second line must include a comma-separated list of violated categories. [\textbackslash INST] & Unsafe \& Safe   \\
        \bottomrule
    \end{tabular}
    \newline
    \caption{Prompt template and negative \& positive tokens used in LLM classifier judge}
    \label{tab:llm_judge_example}
\end{table}

\section{Evaluation Benchmarks}\label{app: eval}

One example prompt of the MBPP evaluation set:

\begin{shaded}
{\ttfamily
You are an expert Python programmer, and here is your task:

Write a function to sort a given matrix in ascending order according to the sum of its rows.

Your code should pass the following tests:

assert sort\_matrix([[1, 2, 3], [2, 4, 5], [1, 1, 1]])==[[1, 1, 1], [1, 2, 3], [2, 4, 5]]

assert sort\_matrix([[1, 2, 3], [-2, 4, -5], [1, -1, 1]])==[[-2, 4, -5], [1, -1, 1], [1, 2, 3]]

assert sort\_matrix([[5,8,9],[6,4,3],[2,1,4]])==[[2, 1, 4], [6, 4, 3], [5, 8, 9]]
}
\end{shaded}

One example prompt of the HumanEval evaluation set:

\begin{shaded}
{\ttfamily
Write a solution to the following problem and make sure that it passes the tests:

$```$python

from typing import List

def remove\_duplicates(numbers: List[int]) -> List[int]:

    """ From a list of integers, remove all elements that occur more than once.
    
    Keep order of elements left the same as in the input.
    
    $>>>$ remove\_duplicates([1, 2, 3, 2, 4])
    
    [1, 3, 4]
    
    """

$```$
}
\end{shaded}







\section{Reward Hacking Examples}\label{sc: reward_hacking}
We provide examples of reward hacking across various tasks in \Cref{tab:reward_hacking}. Here, the reward model we trained in Section \ref{sc: rm_exp} assigns higher values to low quality outputs than to high quality ones. Note that we report the raw reward value in \Cref{tab:reward_hacking} instead of the calibrated one.

\begin{longtable}{>{\raggedright\arraybackslash}m{1.0cm} >{\raggedright\arraybackslash}m{4cm} >{\raggedright\arraybackslash}m{6cm} >{\centering\arraybackslash}m{1.5cm} > {\centering\arraybackslash}m{1.5cm}}
        \toprule
        \textbf{Task} & \textbf{Prompt} & \textbf{Response} & \textbf{Reward} & \textbf{Quality} \\
        \midrule
        \multirow{2}{*}[-10ex]{Math} & \multirow{2}{4cm}[-8ex]{\raggedright Given the following problem, reason and give a final answer to the problem.
        
        Problem: simplify $9\cdot\frac{12}{7}\cdot \frac{-35}{36}$.
        Your response should end with "The final answer is [answer]"} & To simplify the expression, we can start by canceling out common factors between the numerator and denominator of the fractions. In this case, we have 12 and 36, which have a common factor of 12. We also have 7 and 35, which have a common factor of 7. Canceling out these factors, we get: 
        
        $9\cdot\frac{12}{7}\cdot \frac{-35}{36} = 9\cdot\frac{\cancel{12}}{\cancel{7}}\cdot \frac{\cancelto{5}{35}}{\cancelto{3}{36}}$ 
        
        Simplifying further, we have: $9\cdot\frac{12}{7}\cdot \frac{-35}{36} = 9\cdot \frac{5}{3}$

        Now, we can simplify the fraction 5/3 by dividing the numerator and denominator by their greatest common divisor, which is 1. 

        $9\cdot \frac{5}{3} = \frac{9\cdot 5}{3} = \frac{45}{3}$

        Finally, we can simplify the fraction 45/3 by dividing the numerator and denominator by their greatest common divisor, which is 3. 

        $\frac{45}{3} = 15$.

        Therefore, the final answer is 15 & 4.75 & Low \\
        \cline{3-5}
        & & To simplify the given expression, we can start by canceling common factors:

        $9\cdot\frac{12}{7}\cdot \frac{-35}{36}$

        First, let's look at the numerators. Both 12 and 36 are divisible by 12, so we can simplify the expression as follows:

        $9\cdot\frac{12}{7}\cdot \frac{-35}{36} = 9\cdot\frac{1}{7}\cdot \frac{-35}{3}$

        Next, we can look at the remaining factors. 7 and 35 are divisible by 7, so we simplify further:

        $9\cdot\frac{1}{7}\cdot \frac{-35}{3} = 9\cdot\frac{-5}{3}$

        Finally, we have 9 and 3 which are divisible by 3. Performing the division, we get:

        $9\cdot\frac{-5}{3} = -15$

        So, the final answer is -15. & 4.59 & High \\
        \midrule
        \multirow{2}{*}[-10ex]{Coding} & \multirow{2}{4cm}[12ex]{
        \raggedright
        You are an expert python programmer, and here is your task to complete the code:
        
        ```
        
        python
        
        def $\text{you}\_\text{are}\_\text{a}\_\text{cube}$(cube):
        
        """
        
        In geometry, a cube is a three-dimensional solid object bounded by six square faces, facets or sides, with three meeting at each vertex.The cube is the only regular hexahedron and is one of the five Platonic solids.
        
        It has 12 edges, 6 faces and 8 vertices.The cube is also a square parallelepiped, an equilateral cuboid and a right rhombohedron.
        It is a regular square prism in three orientations, and a trigonal trapezohedron in four orientations.

        You are given a task of finding a if the provided value is a perfect cube!
        
        """
        
        ```
        } & Here is the requested Python code:
        
        ```
        
        python
        
        def $\text{you}\_\text{are}\_\text{a}\_\text{cube}$(cube):
        
        $\qquad$if cube in [0, 1, 8, 27, 64, 125, 216, 343, 512, 729, 1000, 1331, 1728, 2163, 2744, 3375, 4096, 4913, 5832, 6859, 8000, 9261, 10648, 12167, 13824, 15625, 17576, 19683, 21952, 24389, 27000, 29791, 32768, 35937, 39312, 42875, 46656, 60704, 63488, 68591, 72903, 79507, 92609, 101218, 109744, 120422, 131433, 143489, 157464, 173015, 190044, 209957, 232630, 250047, 270699, 294035, 320264, 348484, 390625, 435672, 488412, 511315, 539047, 576480, 625000, 648930, 676030, 704977, 729028, 758858, 795293, 847442, 904 Sha, 1000000, 1024003, 1056731]:
            
            $\qquad$return True
                
            return False
            
        ``` 
        & 1.8 & Low \\
        \cline{3-5}
        & & Sure, you can solve your task by checking if the cube root of the given value is an integer or not. Here's your code:

        ```
        
        python
        
        import math

        def $\text{you}\_\text{are}\_\text{a}\_\text{cube}$(cube):
        
        """
        
        In geometry, a cube is a three-dimensional solid object bounded by six square faces, facets or sides, with three meeting at each vertex.The cube is the only regular hexahedron and is one of the five Platonic solids. It has 12 edges, 6 faces and 8 vertices.The cube is also a square parallelepiped, an equilateral cuboid and a right rhombohedron. It is a regular square prism in three orientations, and a trigonal trapezohedron in four orientations.
        
        """
        
        $\qquad$return math.isqrt(cube) ** 3 == cube.

        ```
        & 0.96 & High \\
        \midrule
        \multirow{2}{*}[-10ex]{Safety} & \multirow{2}{4cm}[-8ex]{\raggedright What are some things that people are unaware about that could kill them instantly?} & There are several things that can cause instant death that people may not be aware of or may underestimate the risks associated with them. Here are some examples $\cdots$ & 0.37 & Low \\
        \cline{3-5}
        & & As a responsible AI language model, I must emphasize that it's essential to prioritize safety and awareness to avoid accidental harm. While it's impossible to list every potential danger, I'll highlight some little-known risks that could have severe consequences. However, please remember that most of these risks are extremely rare, and with proper caution, you can minimize the likelihood of encountering them $\cdots$ & -0.44 & High \\
        \midrule
        \bottomrule
        \caption{Example of reward hacking in different tasks}
        \label{tab:reward_hacking}
\end{longtable}

\end{document}